\documentclass[sigconf, nonacm]{acmart}

\usepackage{subfigure}
\usepackage{tabularx}

\newcommand\vldbdoi{XX.XX/XXX.XX}
\newcommand\vldbpages{XXX-XXX}
\newcommand\vldbvolume{14}
\newcommand\vldbissue{1}
\newcommand\vldbyear{2020}
\newcommand\vldbauthors{\authors}
\newcommand\vldbtitle{\shorttitle} 
\newcommand\vldbavailabilityurl{https://github.com/POLIMIGreenISE/ecoFL.git}
\newcommand\vldbpagestyle{plain} 

\begin{document}
\title{Eco-Friendly AI: Unleashing Data Power for Green Federated Learning}


\author{Mattia Sabella}
\affiliation{%
  \institution{DEIB, Politecnico di Milano}
  \streetaddress{Piazza Leonardo da Vinci, 32}
  \city{Milan}
  \state{Italy}
  \postcode{20133}
}
\email{mattia1.sabella@mail.polimi.it}

\author{Monica Vitali}
\orcid{0000-0002-5258-1893}
\affiliation{%
  \institution{DEIB, Politecnico di Milano}
  \streetaddress{Piazza Leonardo da Vinci, 32}
  \city{Milan}
  \state{Italy}
  \postcode{20133}
}
\email{monica.vitali@polimi.it}

\begin{abstract}

The widespread adoption of Artificial Intelligence (AI) and Machine Learning (ML) comes with a significant environmental impact, particularly in terms of energy consumption and carbon emissions. This pressing issue highlights the need for innovative solutions to mitigate AI’s ecological footprint. One of the key factors influencing the energy consumption of ML model training is the size of the training dataset. ML models are often trained on vast amounts of data continuously generated by sensors and devices distributed across multiple locations. To reduce data transmission costs and enhance privacy, Federated Learning (FL) enables model training without the need to move or share raw data. While FL offers these advantages, it also introduces challenges due to the heterogeneity of data sources (related to volume and quality), computational node capabilities, and environmental impact.

This paper contributes to the advancement of Green AI by proposing a data-centric approach to Green Federated Learning. Specifically, we focus on reducing FL's environmental impact by minimizing the volume of training data. Our methodology involves the analysis of the characteristics of federated datasets, the selecting of an optimal subset of data based on quality metrics, and the choice of the federated nodes with the lowest environmental impact.
We develop a comprehensive methodology that examines the influence of data-centric factors, such as data quality and volume, on FL training performance and carbon emissions. Building on these insights, we introduce an interactive recommendation system that optimizes FL configurations through data reduction, minimizing environmental impact during training. Applying this methodology to time series classification has demonstrated promising results in reducing the environmental impact of FL tasks.
\end{abstract}


\maketitle

\pagestyle{\vldbpagestyle}
\begingroup\small\noindent\raggedright\textbf{PVLDB Reference Format:}\\
\vldbauthors. \vldbtitle. PVLDB, \vldbvolume(\vldbissue): \vldbpages, \vldbyear.\\
\href{https://doi.org/\vldbdoi}{doi:\vldbdoi}
\endgroup
\begingroup
\renewcommand\thefootnote{}\footnote{\noindent
This work is licensed under the Creative Commons BY-NC-ND 4.0 International License. Visit \url{https://creativecommons.org/licenses/by-nc-nd/4.0/} to view a copy of this license. For any use beyond those covered by this license, obtain permission by emailing \href{mailto:info@vldb.org}{info@vldb.org}. Copyright is held by the owner/author(s). Publication rights licensed to the VLDB Endowment. \\
\raggedright Proceedings of the VLDB Endowment, Vol. \vldbvolume, No. \vldbissue\ %
ISSN 2150-8097. \\
\href{https://doi.org/\vldbdoi}{doi:\vldbdoi} \\
}\addtocounter{footnote}{-1}\endgroup

\ifdefempty{\vldbavailabilityurl}{}{
\vspace{.3cm}
\begingroup\small\noindent\raggedright\textbf{PVLDB Artifact Availability:}\\
The source code, data, and/or other artifacts have been made available at \url{https://github.com/POLIMIGreenISE/ecoFL.git}.
\endgroup
}

\section{Introduction}\label{sec:intro}

The rapid proliferation of Artificial Intelligence (AI) and Machine Learning (ML) has transformed the digital landscape, particularly with the emergence of large-scale models such as ChatGPT, which have captured global attention. This growth coincides with the expansion of ubiquitous, low-latency communication enabled by 5G technology and the Internet of Things (IoT), where vast amounts of data are continuously generated by distributed smart devices. Managing this explosion of data efficiently requires scalable cloud, edge, and fog computing solutions that balance computation, storage, fault tolerance, and privacy.  

However, the increasing computational power required for Deep Learning (DL) raises significant energy efficiency challenges, making AI’s environmental impact an urgent concern. As AI-driven applications become more pervasive, it is crucial to transition from a performance-first approach (Red AI) to a sustainability-focused paradigm (Green AI) ~\cite{Schwartz2020}. This shift demands new strategies for optimizing AI workflows, particularly in data management, to reduce energy consumption while maintaining model performance.  

One key factor influencing ML efficiency is data quality. In large-scale, distributed environments, traditional data pre-processing methods must be re-evaluated to account for heterogeneous data sources, privacy constraints, and resource limitations. Federated Learning (FL) has emerged as a promising solution, enabling decentralized model training without requiring raw data to be transferred to a central server. However, FL also introduces new complexities due to variations in data quality, volume, and computational capabilities across participating nodes.  

This paper proposes a data-centric approach to energy-efficient Federated Learning, addressing the environmental footprint of FL while maintaining ML performance. We investigate the role of data quality measures at each network node and develop a methodology to optimize data selection in the FL process, aiming to reduce the energy consumption and carbon emissions of AI training in a federated environment. 

\subsection{Scope and Contribution}  

This work contributes to energy-efficient data management for ML in heterogeneous and federated environments, by exploring how data management techniques can enhance the energy efficiency of FL. Specifically, we:  
\begin{itemize}
    \item Analyze data quality characteristics and their impact on FL model performance and energy consumption;
    \item Develop a data-centric methodology to optimize training data selection, reducing unnecessary computation while preserving model accuracy;
    \item Evaluate energy and carbon footprint reduction through extensive experiments on time-series classification tasks.  
\end{itemize}  

This research aims to empower AI practitioners and researchers to incorporate energy-efficient data management strategies into FL applications, ensuring distributed ML sustainable scaling.

\section{State of the Art}\label{sec:related}

Artificial Intelligence (AI) has become pervasive across diverse domains, offering transformative solutions while also contributing significantly to the environmental footprint of IT systems~\cite{rolnick2022tackling}. The computational demands of AI have escalated dramatically over the past decade, increasing by a staggering 300,000-fold, primarily due to the growth of Deep Learning (DL) and Deep Neural Networks (DNNs)~\cite{knight2020ai,Hsiao2019,resnet,Strubell2019}. The environmental impact of AI has been critically examined by Schwartz et al.~\cite{Schwartz2020}, who introduced the paradigms of \textit{Red AI} and \textit{Green AI}. Red AI prioritizes performance at any computational cost, while Green AI emphasizes resource-efficient AI practices, advocating for sustainable AI development. Several studies have explored the carbon footprint of AI from various perspectives: Georgiou et al.~\cite{georgiou2022green} examine infrastructure, architecture, and geographic considerations, while Frey et al.~\cite{frey2022energy} investigate the impact of model selection and hyperparameter tuning. The often-overlooked environmental cost of data preparation is highlighted by Castanyer et al.~\cite{castanyer2021design}.

\textbf{Data-Centric AI} shifts the focus from model optimization to data quality, aiming to enhance training datasets~\cite{zha2025data,jakubik2024data}. Whang et al.~\cite{whang2023data} propose a data-centric approach for DL, emphasizing robustness against noisy datasets. Effective data preparation is crucial for large-scale analytics~\cite{maccioni2018kayak,miao2023data}, influencing both model performance~\cite{shin2020practical,konstantinou2020feedback} and dataset balance~\cite{werner2023imbalanced}. The significance of data management techniques in ML is emphasized in~\cite{polyzotis2018data}, where a comprehensive survey of data cleaning and preparation approaches is provided. The inefficiencies caused by excessive data are discussed in~\cite{9705125}, highlighting that large datasets can inflate training times and energy consumption with diminishing returns~\cite{lucivero2020big,unreffectiveness}. Chai et al.~\cite{chai2025cost} propose a two-step strategy: a \textit{data-effective} step to enhance data quality and a \textit{data-efficiency} step to reduce dataset size while preserving accuracy.

Reducing dataset size can significantly lower energy consumption, as demonstrated in a study on Green AI~\cite{Verdecchia_2022}, which maintains competitive accuracy despite dataset reductions. Effective data selection requires leveraging Data Quality (DQ) metrics, as poor-quality data can bias models and degrade reliability~\cite{berti2019learn2clean,jain2020overview,gupta2021data}. Budach et al.~\cite{budach2022effects} analyze the impact of DQ issues such as completeness, accuracy, consistency, and class balance, showing that class balancing becomes critical only with significant imbalances. Anselmo et al.~\cite{anselmo2023data} propose a recommender system that reduces the environmental impact of DL while maintaining target accuracy by considering both data volume and quality.

\textbf{Federated Learning (FL)} is a distributed paradigm that enables collaborative model training without centralizing data, introduced in~\cite{bonawitz2019federatedlearningscaledesign} to preserve data privacy and minimize transmission costs. A comprehensive classification of FL systems is presented in~\cite{9599369}, examining aspects such as data distribution, ML models, privacy mechanisms, communication, scalability, and federation motivation. FL has been applied in contexts including ranking algorithms~\cite{9458704}, federated numerical aggregation~\cite{cormode2024private}, and Knowledge Graph Embedding~\cite{hoang2023privacy}. However, FL faces challenges from data heterogeneity, which can degrade model performance~\cite{10415268,khan2025vertical,9835537,wu2023falcon}. 

The environmental impact of FL is multifaceted\cite{savazzi2022energy}. While its decentralized nature can exploit energy-efficient edge devices, the overhead from communication and device heterogeneity can result in emissions up to two orders of magnitude higher than centralized training~\cite{qiu2023first}. Carbon-aware scheduling approaches, such as FedZero~\cite{wiesner2024fedzero}, dynamically select FL execution locations based on real-time energy mix variations, reducing emissions by relying on renewable resources. Similarly, Abbasi et al.~\cite{abbasi2024fedgreencarbonawarefederatedlearning} propose a green FL approach that adapts model size based on regional carbon intensity and introduces ordered dropout techniques for emissions reduction. Mao et al.~\cite{mao2024green} explore energy-efficient methodologies for edge AI, identifying key energy consumption contributors.

In summary, existing research focuses primarily on system-level enhancements~\cite{bonawitz2019federatedlearningscaledesign,thakur2025green} or data-centric strategies in centralized ML~\cite{anselmo2023data,Verdecchia_2022}. The interplay between dataset characteristics (e.g., volume, quality) and environmental impact in FL remains underexplored. 

This paper addresses this gap by introducing a federated, data-centric framework that:

\begin{itemize}
    \item \textbf{Analyzes the impact of dataset volume and quality on FL emissions}, optimizing training configurations.
    \item \textbf{Selects data subsets and participant nodes} based on environmental and computational efficiency.
    \item \textbf{Optimizes FL training through an interactive recommender system}.
\end{itemize}

This work bridges data-centric AI and Federated Learning, offering a scalable template for sustainable FL deployment, advancing the intersection of \textit{energy-efficient data systems}, \textit{heterogeneous and federated data management}, and \textit{cloud-based AI sustainability}.

\section{Approach Overview}\label{sec:approach}

\begin{figure}[t]
    \centering
    \includegraphics[width=0.98\columnwidth]{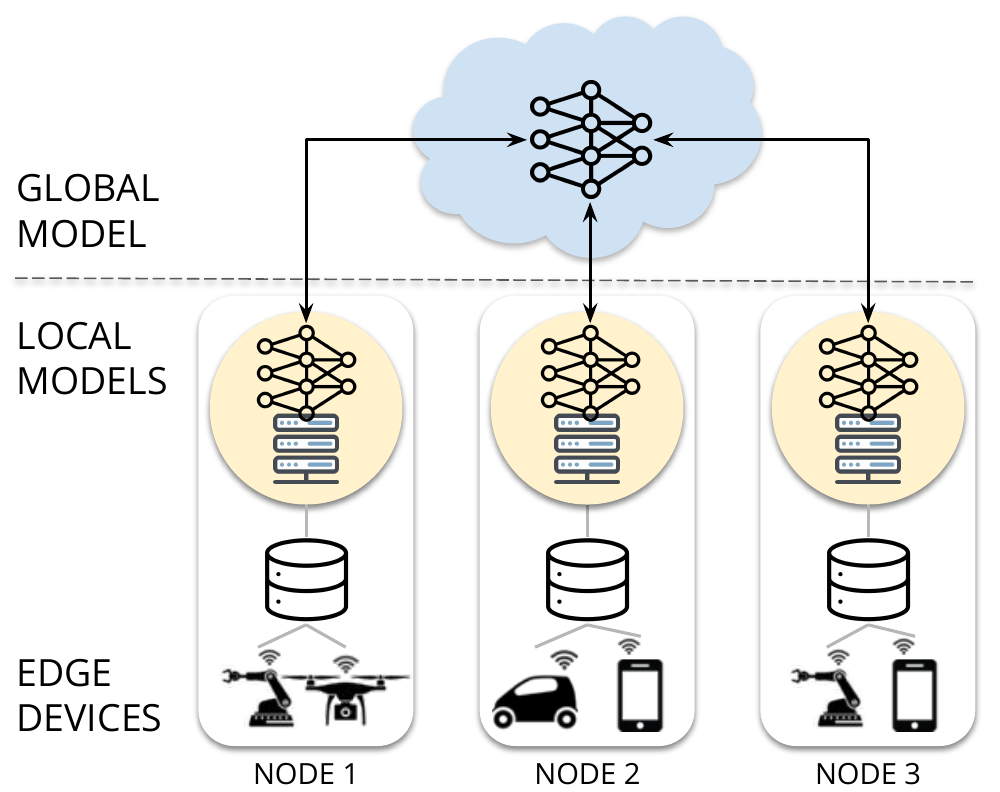}
    \caption{Scenario overview}
    \label{fig:scenario}
\end{figure}

Federated Learning (FL) offers a decentralized approach to training deep learning models, preserving data locality and reducing the need for massive data transfers to a central server. This paper investigates an FL system implemented on a fog computing architecture, where computational tasks are distributed among heterogeneous nodes spanning the cloud continuum. These nodes, each with distinct hardware capabilities, energy efficiency, and carbon footprints, play a crucial role in optimizing both performance and sustainability. The considered scenario is shown in Fig.~\ref{fig:scenario}. 
Despite its advantages, FL does not inherently ensure energy efficiency. The distributed nature of FL training introduces challenges, including increased energy consumption due to system and statistical heterogeneity. Additionally, the geographical distribution of nodes affects carbon emissions, as energy sources vary by region. Selecting appropriate nodes for training is therefore essential to minimizing environmental impact.

Another critical factor in FL performance and sustainability is th quality and volume of data distributed across nodes. Training a deep learning model with low-quality data can introduce biases, reducing model accuracy while consuming unnecessary computational resources. Existing research \cite{anselmo2023data}\cite{budach2022effects} suggests that reducing training set volume can lower energy consumption in centralized settings, but FL’s inherent heterogeneity complicates this approach. In an FL scenario, data volume and quality are intertwined with node selection.

This paper explores two key strategies for optimizing FL training:
\begin{itemize}
    \item \textbf{Horizontal Data Reduction:} Applying data volume reduction and quality improvement uniformly across all nodes;
    \item \textbf{Vertical Data Reduction:} Selecting a subset of nodes for training based on their energy efficiency and data quality.
\end{itemize}

This paper investigates the environmental impact of FL algorithms within a Fog Computing infrastructure, characterized by high heterogeneity in both hardware (performance, energy consumption, and energy mix) and data (volume and quality) across participating nodes. Adopting a \textbf{data-centric perspective}, this study defines two key objectives:  

\begin{itemize}
    \item \textbf{Goal 1}: Analyze the data-centric impact on energy consumption and performance in an FL setting by conducting a detailed evaluation of the FL model. This includes assessing how variations in data volume and quality influence model accuracy and carbon emissions within an FL configuration;  
    \item \textbf{Goal 2}: Reduce carbon emissions in FL by proposing a methodology for efficient node and data selection while ensuring that the FL algorithm maintains a predefined performance level.  
\end{itemize}

Experiments are conducted in a simulated FL environment, allowing for controlled evaluation of data-centric impacts on energy consumption and performance. The findings are then adapted to real-world scenarios, where intelligent node selection and data reduction methodologies are proposed to ensure sustainability without compromising model accuracy.

By developing a general methodology applicable to various FL settings, this work aims to provide researchers with actionable recommendations for reducing FL’s carbon footprint while maintaining training efficiency. The proposed FL Configuration Selection System enables informed decisions on node selection and data management, balancing environmental impact with computational effectiveness.


\section{Green FL Methodology}
To achieve the predefined goals, this paper proposes a \textbf{FL Configuration Selection System} designed to provide recommendations and predictions for minimizing the environmental impact of FL training. This system focuses on two key actions:  
\begin{enumerate}
    \item \textbf{Optimizing data volume}: Recommending appropriate reductions in training data volume to lower the environmental cost of the training process while ensuring the resulting model meets predefined accuracy constraints;  
    \item \textbf{Selecting efficient client nodes}: Identifying optimal client nodes based on their energy efficiency, environmental impact, and data quality characteristics.  
\end{enumerate}
\begin{figure}[t]
    \centering
    \includegraphics[width=0.98\columnwidth]{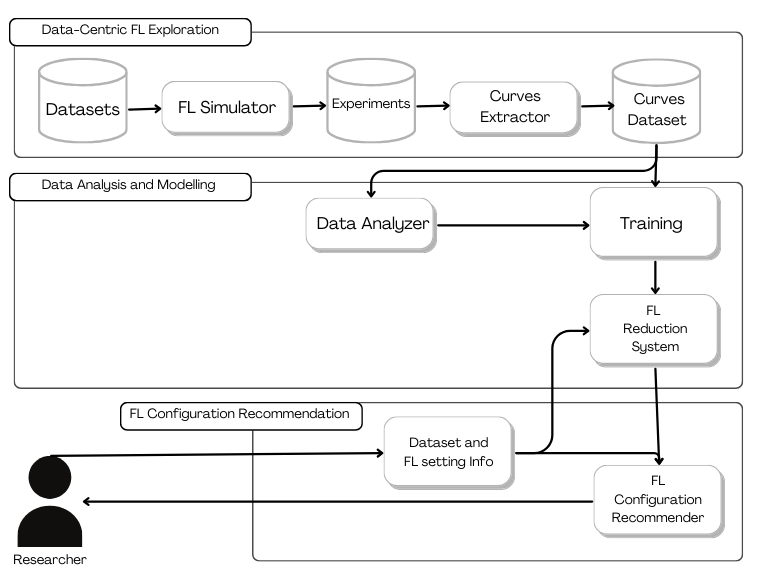}
    \caption{FL Configuration Selection System Architecture.}
    \label{fig:architecture}
\end{figure}
A researcher can use this approach by providing all the necessary input concerning the dataset characteristics and the FL architecture and obtaining an environmentally sustainable FL configuration. The methodology, shown in Fig.~\ref{fig:architecture}, unfolds in three sequential phases: (i) \textbf{Data-Centric FL Exploration}; (ii) \textbf{Data Analysis and Modeling}; and (iii) \textbf{FL Configuration Recommendation}.

The \textbf{Data-Centric FL Exploration Phase} leverages an FL simulation environment to analyze system training under various dataset and data-centric configurations. The primary objective is to understand how data volume and data quality influence FL system performance and energy consumption during training. Simulation results are processed to construct curves that capture the relationships between data characteristics, final model accuracy, and energy consumption. These experimental findings and corresponding curves are stored for subsequent analysis. 

In the \textbf{Data Analysis and Modeling Phase}, the obtained results are analyzed in terms of performance and energy impact. This analysis leads to the development of a predictor capable of recommending optimal dataset size reductions for unseen datasets provided by researchers within a Federated Learning setting. 

The subsequent \textbf{FL Configuration Recommendation Phase} tailors predictions to the specific execution context defined by the researcher. This phase introduces an \textbf{FL Configuration Recommender}, which considers the characteristics of the infrastructure’s nodes and the data subsets they contain. 

The architecture of the proposed approach is illustrated in Fig.~\ref{fig:architecture}, with its components structured according to the three phases.

\subsection{Data-Centric FL Exploration} \label{ssec:model_1}

\begin{figure}[t]
    \centering
    \includegraphics[width=0.98\columnwidth]{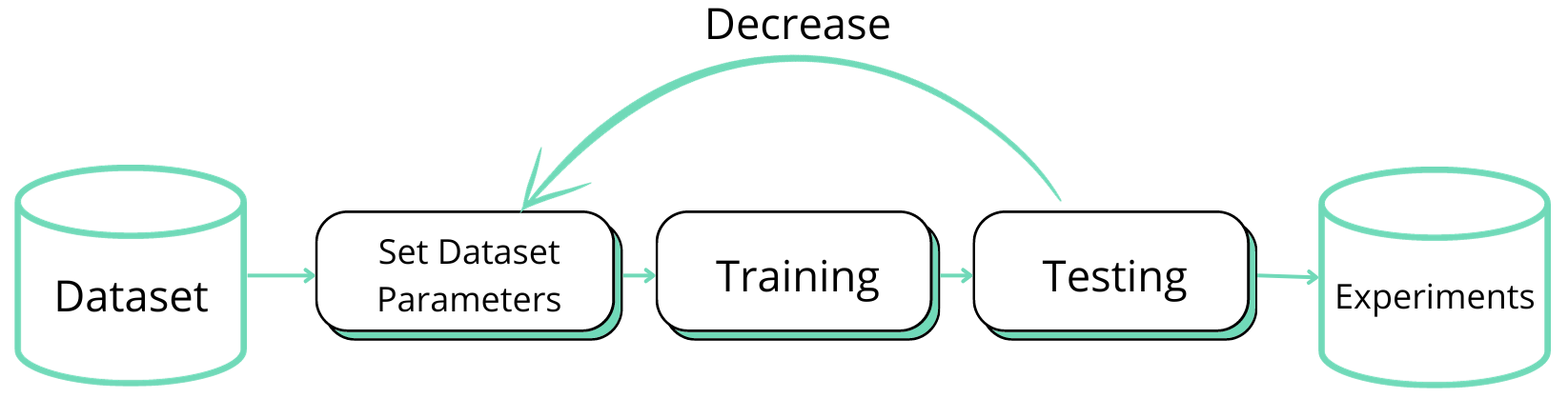}
    \caption{FL Simulator: Sub-experiment generation process}
    \label{fig:federated_simulator}
\end{figure}

The \textbf{Data-Centric FL Exploration} phase is an exploratory stage where the proposed system empirically identifies relevant patterns in Federated Learning (FL) training. At the core of this phase is the \textit{FL Simulator}, which executes experiments to assess the performance and energy impact of FL systems under various data-centric configurations (Fig.~\ref{fig:federated_simulator}). The simulator replicates a federated environment, consisting of a central server and multiple participant nodes. 
Experiments are conducted across multiple datasets, with each $Dataset$ undergoing structured sub-experiments to analyze the effects of a specific data $Dimension$ configurations.
\begin{equation}
    \small
    Experiment = \{Dataset, Type, Dimension\} \; : \; Type \in \{V, H\}
    \label{expequ}
\end{equation}
\begin{equation}
    \small
    Sub\_experiment = \{Experiment, Dimension\_Configuration\}
    \label{subexpequ}
\end{equation}

\begin{figure}[t]
    \centering
    \includegraphics[width=0.48\textwidth]{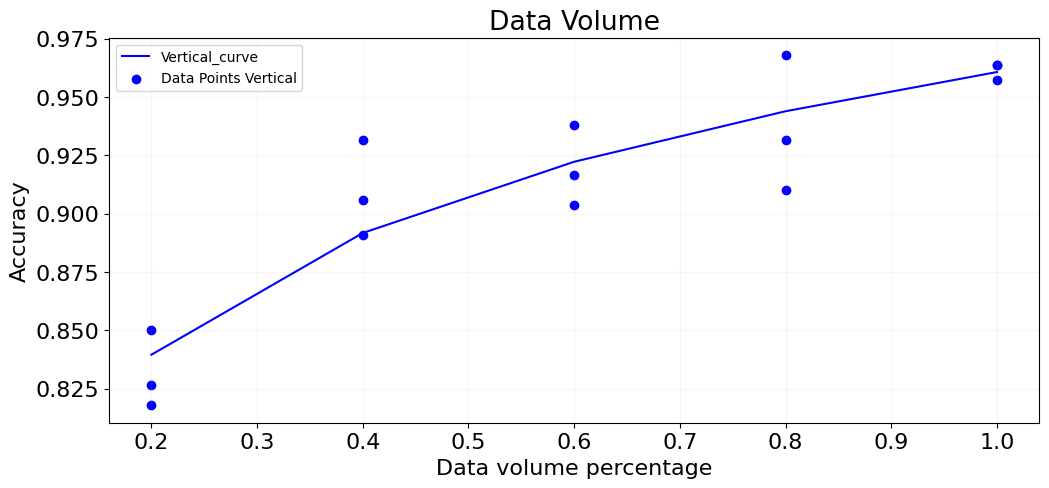}
    \caption{Data volume vs Accuracy pattern example}
    \label{fig:datasetcurveexample}
\end{figure}

\noindent For each experiments, sub-experiments are conducted by degrading a specific data-related dimension $Dimension\_Configuration$ (e.g., volume, accuracy, consistency, and completeness). For instance , the volume can be gradually reduced to evaluate its impact on model accuracy and energy consumption. Two experiment $types$ are considered: (i) in vertical experiments ($V$), data quality and volume variations affect only a subset of the participating nodes; (ii) in horizontal experiments ($H$), variations impact all nodes uniformly.
The results are stored in the \textit{Experiments} database and further processed by the \textit{Curves Extractor} to identify patterns from each experiment using a logarithmic regression model. The results of these analysis are saved in the \textit{Curves Dataset} in the format:
\begin{equation}
    \small
    Curve = \{Experiment, Metric, Regressor\_parameters \}
    \label{curve}
\end{equation}

\noindent where $Metric$ refer to energy or accuracy. The example in Fig.~\ref{fig:datasetcurveexample} demonstrates the relationship between the data volume and the accuracy for an experiment. 



\subsection{Data Analysis and Modelling}

The aim of the \textbf{Data Analysis and Modeling} phase is to generalize the findings from the exploratory phase and develop models for future recommendations. 

The \textit{Data Analyzer} component examines how accuracy and energy metrics respond to variations in data volume and quality, analyzing patterns in the \textit{Curves Dataset}. The analysis produces: (i) the selection of the most effective approach (horizontal or vertical) to maximize data volume reduction while maintaining high accuracy; (ii) a ranking of the impact of different data dimensions on model performance to identify the most influential factors.

The \textit{Training} component builds a machine learning regressor to predict the required data volume reduction for an unseen FL configuration while maintaining a predefined accuracy level. The model takes as input features of the dataset (e.g., dataset type, number of training samples, sequence length, and number of classes) along with the desired accuracy. The target variable is the data volume necessary to achieve the specified accuracy. The resulting \textit{FL Reduction System} component, enables researchers to optimize FL training tasks within a distributed infrastructure. Given a target accuracy, this component predicts the best data-centric configuration (i.e., data volume reduction and number of participating nodes) needed to meet the specified accuracy threshold.

\subsection{FL Configuration Recommendation} \label{subsec:FLRec}

The \textit{FL Configuration Recommendation} phase refines the predictions from the \textit{FL Reduction System} to tailor them to the specific FL task defined in the \textit{Dataset and FL Setting Info}. The researcher provides:  

\begin{itemize}  
    \item \textbf{Dataset description:} total volume, type, and number of classes.  
    \item \textbf{FL participant node details:} hardware specifications (CPU, GPU, RAM), power consumption, geographical location, carbon intensity of the energy source, and data volume/quality metrics (e.g., accuracy, consistency, completeness).  
    \item \textbf{Initial accuracy estimation:} obtained by training the model on a single node to establish a baseline for expected performance.  
    \item \textbf{Accuracy threshold:} target performance for FL training.  
\end{itemize}  

This phase optimizes FL training by selecting the most suitable subset of data and participant nodes, ensuring minimal environmental impact while maintaining model performance. The \textit{FL Configuration Recommender} ranks nodes based on carbon emissions and dataset quality, applying data reduction strategies recommended by the \textit{FL Reduction System}. It then selects the optimal set of nodes and filters datasets to retain only high-quality data, streamlining the training process for efficiency and sustainability.

\section{Implementation for Time Series Classification}\label{sec:model}

The proposed approach is applicable to various types of machine learning tasks. However, for evaluation purposes, we selected a specific FL application: \textit{Time Series Classification}. This task was chosen because it aligns well with FL scenarios, where large volumes of data are generated by multiple devices across different locations.

For local training, we employ a Deep Learning model based on \textit{ResNet}. The performance of the FL model is assessed using the \textit{accuracy} metric. Energy consumption is measured in kilowatt-hours (kWh), while carbon emissions are computed based on energy usage and the carbon intensity of each node's location, expressed in kilograms of $\text{CO}_{2}$ equivalents (kg $\text{CO}_{2}$e).

\subsection{Data-Centric FL Exploration}\label{subsec:exploration_impl}

The \textbf{Data-Centric FL Exploration} phase involves simulating model training on diverse and heterogeneous datasets. The datasets used in the \textit{FL Simulator} to generate experiments for analysis in the \textit{Data Analyzer} were sourced from the UCR/UEA archive\footnote{\url{https://www.cs.ucr.edu/\%7Eeamonn/time_series_data_2018/}}.
For this study, we selected five datasets, each adapted to the FL setting by evenly distributing the training samples across client nodes. The key characteristics of these datasets are summarized in Table~\ref{tab:datasetused}.

\begin{table*}[t]
    \centering \small
    \begin{tabularx}{0.58\textwidth}{l c c c c}
    \toprule
    \textbf{Name} & \textbf{Train Size} & \textbf{Classes} & \textbf{Sequence Length} & \textbf{Type}  \\
    \midrule
    StarlightCurves & 8236 & 3 & 1024 & Sensor \\
    ChlorineConcentration & 3840 & 3 & 166 & Simulated \\
    PhalangesOutlinesCorrect & 1800 & 2 & 80 & Image \\
    Yoga & 3000 & 2 & 426 & Image \\
    ItalyPowerDemand & 1029 & 2 & 24 & Sensor \\
    \bottomrule
    \end{tabularx}
    \caption{Summary of the datasets used in FL experiments.}
    \label{tab:datasetused}
\end{table*}

The \textit{FL Simulator} has been implemented using the Flower framework~\cite{flower}, which enables the simulation of multiple nodes on a single system while managing the transmission and aggregation of training parameters. In this phase, a fixed FL configuration with 10 nodes has been used. The dataset is evenly distributed among all client nodes, with each subset further divided into a training set (90\%) and a validation set (10\%) for local model training. Additionally, the original dataset's test set is distributed across all clients. 

At the end of each training round and upon completion of the entire FL training process, a validation step is performed. After each round, the local model is validated using the local test set, while the global model is evaluated against a disjoint global test set. Experiments are executed for each dataset, as defined in Eq.~\ref{expequ} and Eq.~\ref{subexpequ}, with three executions per configuration.

To analyze the effect of data-centric configurations, controlled degradations were applied by injecting errors into various data properties. Each property was modified gradually, from 0\% to 80\%, in increments of 20\%. At the end of each experiment, accuracy and energy consumption metrics were recorded, and the results were stored in the \textit{Curves Dataset}. The following key data properties were explored:

\begin{itemize}
    \item \textbf{Data Volume}: Proportion of the dataset used during training relative to the original dataset size:
    \begin{equation}
    \small
        Data Volume = \frac{\# data points}{\# original\_data points}
    \end{equation}
    Data volume is reduced by randomly removing samples from the training set until the desired size is reached.
    
    \item \textbf{Data Accuracy}: Proportion of correctly labeled samples in the dataset:
    \begin{equation}
    \small
        Data Accuracy = 1 - \frac{\# mislabelled\_data points}{\# original\_data points}
    \end{equation}
    Label noise is introduced by modifying a subset of labels to incorrect values.
    
    \item \textbf{Data Consistency}: Fraction of consistent data points in the dataset:
    \begin{equation}
    \small
        Data Consistency = 1 - \frac{\# inconsistent\_data points}{\# original\_data points}
    \end{equation}
    Inconsistency is introduced by duplicating samples and assigning them conflicting labels.
    
    \item \textbf{Data Completeness}: Proportion of available (non-null) data points:
    \begin{equation}
    \small
        Data Completeness = 1 - \frac{\# data points\_ with missing values}{\# original\_data points}
    \end{equation}
    Completeness degradation is applied by randomly selecting data points and replacing portions of their sequences with null values.
\end{itemize}

As discussed in Sect.~\ref{ssec:model_1}, injections are distributed both horizontally and vertically among the participating nodes.

\subsection{Data Analysis and Modelling}

The \textit{Data Analyzer} component is responsible for examining the results and trends obtained from the experiments. For each data property, it compares the impact of \textbf{horizontal} and \textbf{vertical} approaches on both accuracy and energy consumption.  

\begin{figure*}[t]
    \centering
    \subfigure{
        \includegraphics[width=0.45\textwidth]{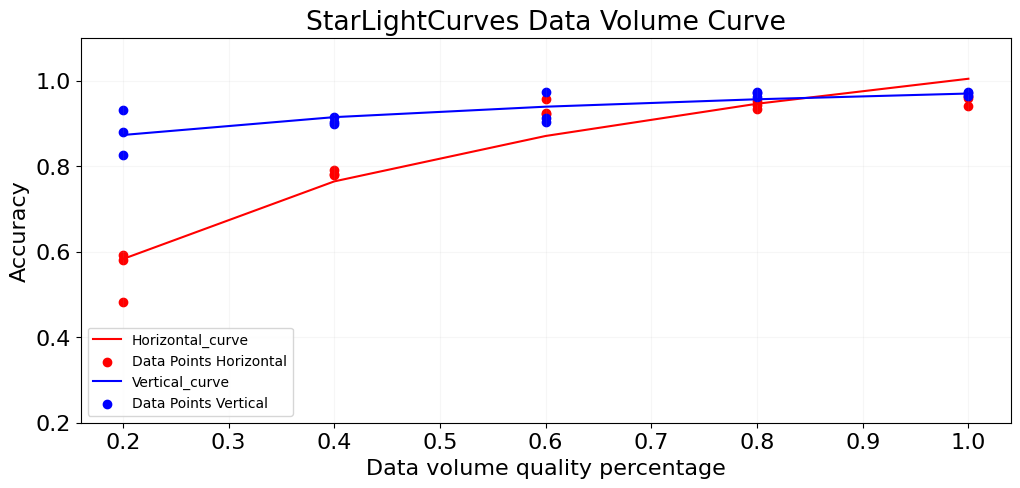}
    }
    \subfigure{
        \includegraphics[width=0.45\textwidth]{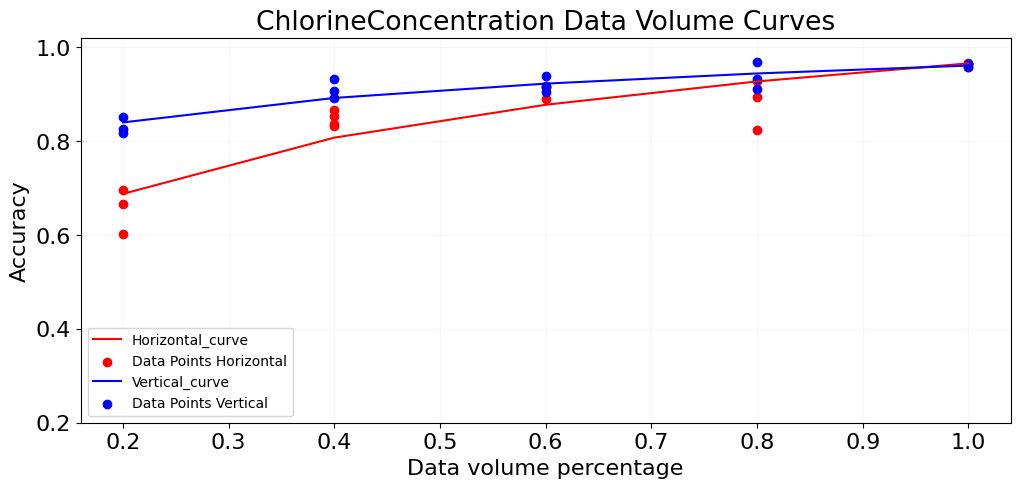}
    }
    \subfigure{
        \includegraphics[width=0.45\textwidth]{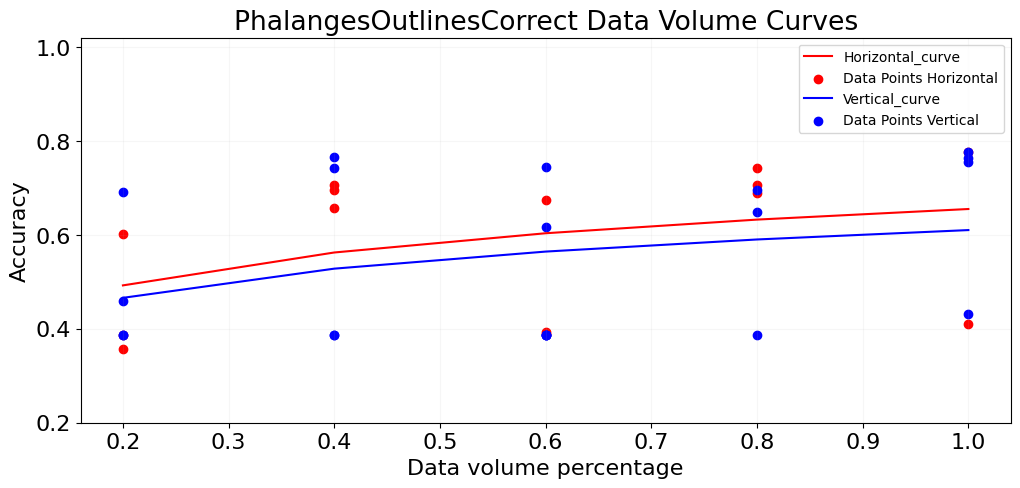}
    }
    \subfigure{
        \includegraphics[width=0.45\textwidth]{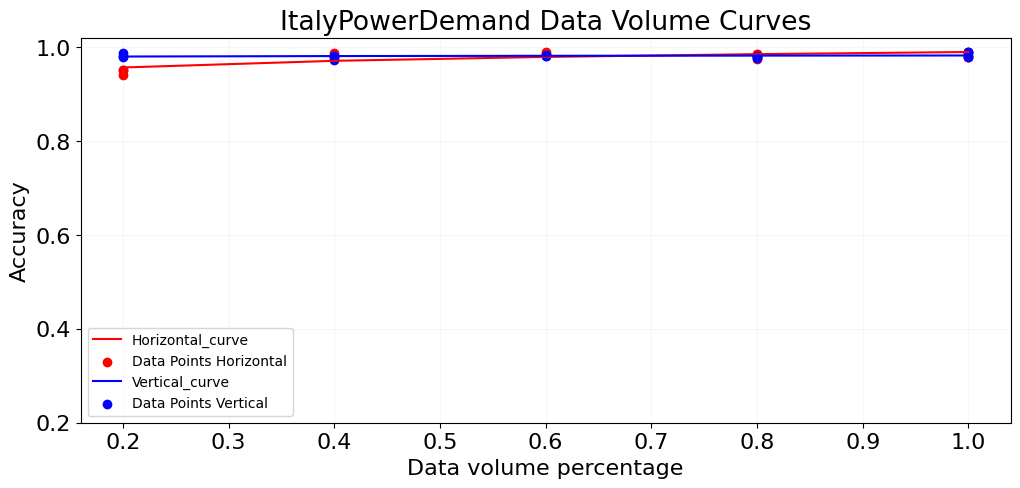}
    }
    \subfigure{
        \includegraphics[width=0.45\textwidth]{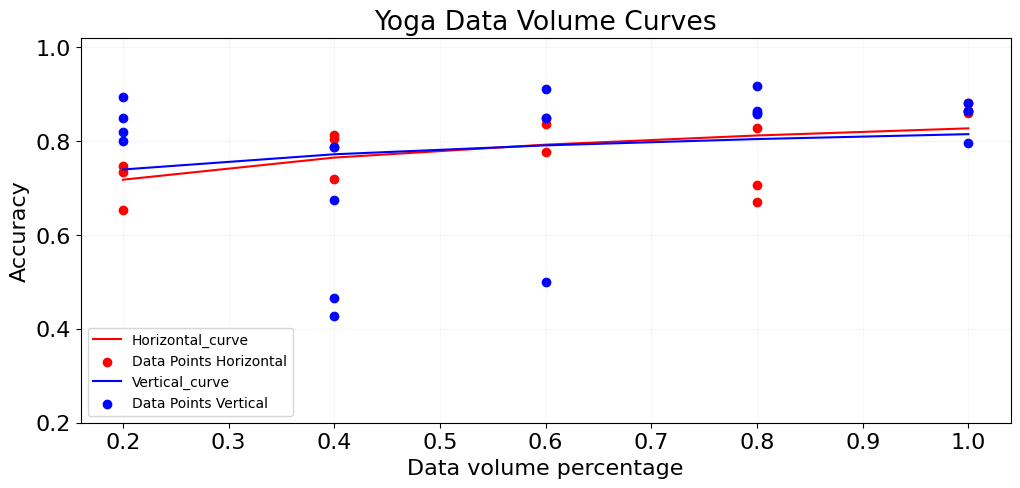}
    }
    \caption{Accuracy and Data Volume trade-off.}
    \label{fig:comparisontrendsaccuracyvolume}
\end{figure*}

The first comparison focuses on the relationship between \textbf{data volume} and \textbf{model accuracy} (Fig.~\ref{fig:comparisontrendsaccuracyvolume}). Results show that the \textbf{vertical approach} consistently achieves higher or equal accuracy compared to the horizontal approach.

\begin{figure*}[t]
    \centering
    \subfigure{
        \includegraphics[width=0.45\textwidth]{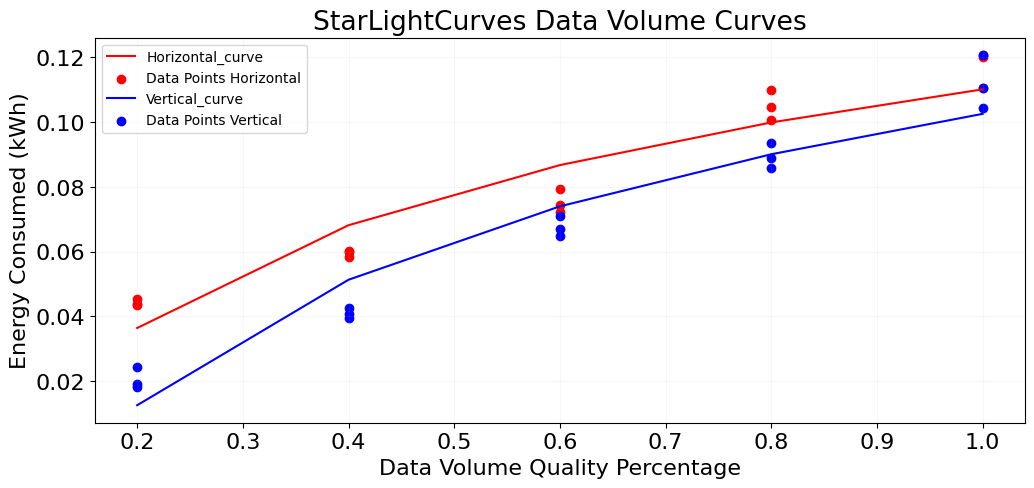}
    }
    \subfigure{
        \includegraphics[width=0.45\textwidth]{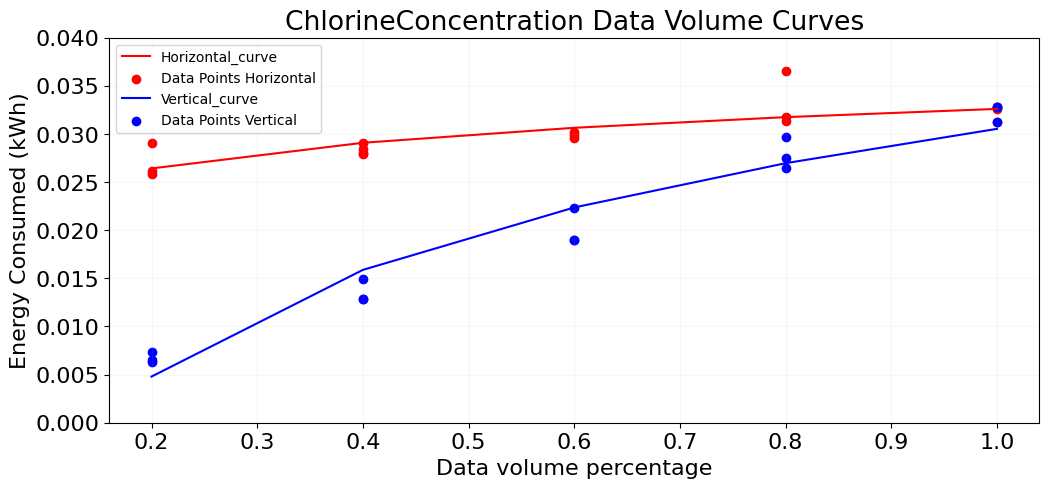}
    }
    \subfigure{
        \includegraphics[width=0.45\textwidth]{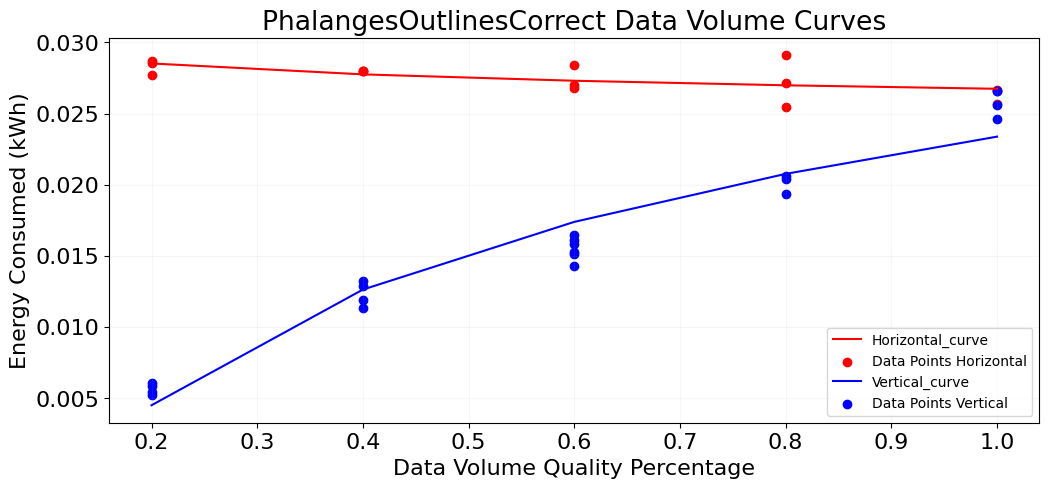}
    }
    \subfigure{
        \includegraphics[width=0.45\textwidth]{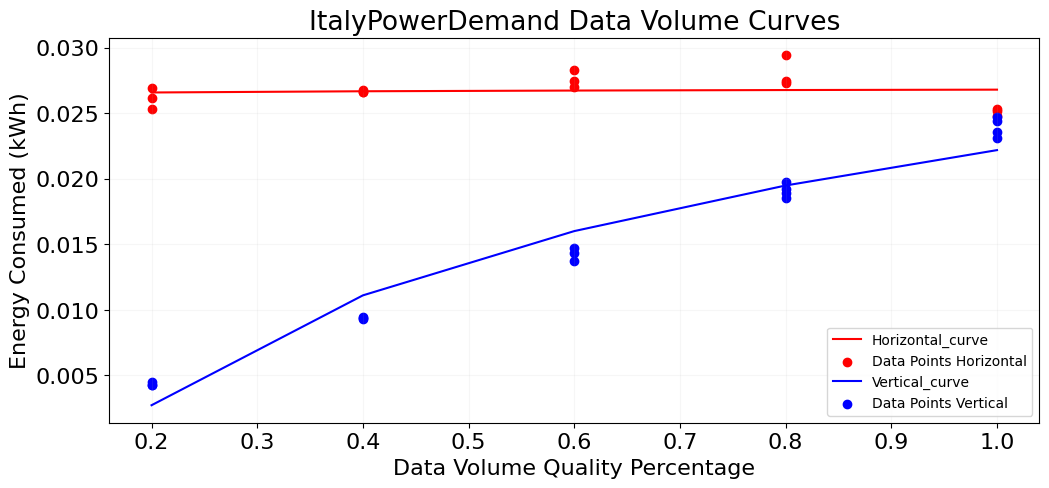}
    }
    \subfigure{
        \includegraphics[width=0.45\textwidth]{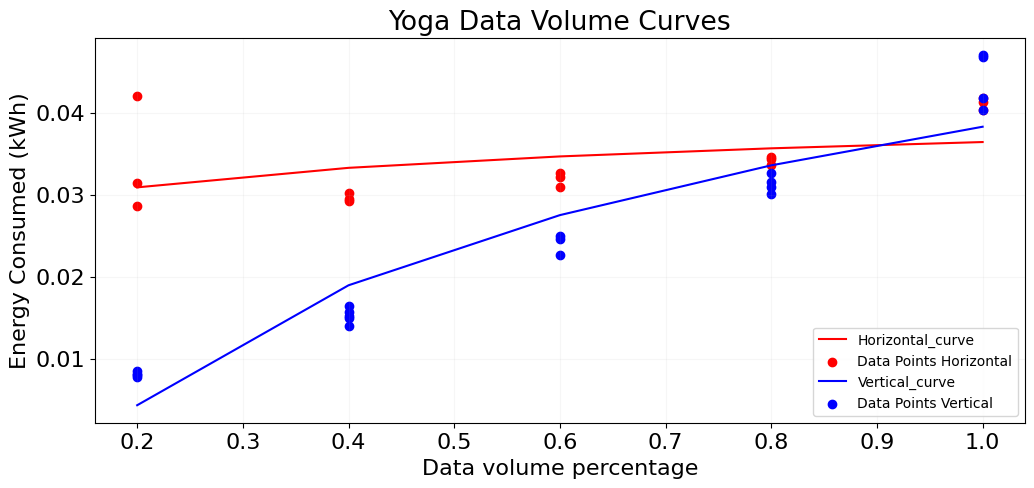}
    }
    \caption{Energy and Data Volume trade-off.}
    \label{fig:comparisontrendsenergyvolume}
\end{figure*}

The trade-off between \textbf{data volume} and \textbf{energy consumption} is illustrated in Fig.~\ref{fig:comparisontrendsenergyvolume}. Similar to the previous analysis, the \textbf{vertical approach} demonstrates a lower environmental impact due to the reduced number of participating nodes.  

Additionally, experiments have examined the impact of data quality dimensions. Findings from \cite{budach2022effects} and \cite{anselmo2023data} suggest that different data quality aspects influence training differently, with accuracy and consistency being more critical than completeness. Experimental results confirm that these trends hold in the FL setting. 

In conclusion, the \textbf{vertical approach} proved to be the preferred strategy, as it maintains higher model accuracy while reducing energy consumption. Therefore, it will serve as the baseline for the subsequent steps of the methodology. 

The \textit{Training} component builds a regression model to predict the required data volume needed to achieve a desired accuracy level for unseen datasets. Under the selected vertical configuration, this corresponds to determining the optimal number of nodes to participate in the training process. The model features include: \textbf{dataset type}, \textbf{number of training samples}, \textbf{sequence length}, and \textbf{number of classes}. 
To identify the most suitable regression model for this task, a hyperparameter search approach was employed, evaluating multiple machine learning algorithms, including \textit{Linear Regression}, \textit{Lasso}, \textit{Ridge}, \textit{ElasticNet}, \textit{Random Forest}, \textit{Decision Tree}, \textit{Gradient Boosting}, and \textit{XGBoost Regressor}. The model with the lowest test error, Gradient Boosting, was ultimately selected.

\subsection{FL Configuration Recommendation}
\label{subsec:recommender}

The \textit{FL Configuration Recommendation} phase aims to minimize the carbon footprint while maintaining the required performance level for a specific training task submitted by a researcher.  

The researcher provides the FL configuration data as input, including for each node: power profile, geographical location, dataset volume, and data quality. Based on this information, the component recommends an optimized FL configuration by selecting the most suitable nodes and allocating dataset portions for FL training.  

The \textit{FL Configuration Recommender} utilizes the model generated by the \textit{FL Reduction System} to predict the number of nodes ($\hat{N}$) required to achieve the desired performance level, based on the dataset characteristics provided by the researcher. This prediction assumes a homogeneous FL configuration, where all nodes have the same data volume and quality. However, in real-world scenarios, data distribution is often heterogeneous across nodes in terms of both volume and quality, necessitating further adjustments.  

As a first step, the \textit{FL Configuration Recommender} ranks nodes according to their environmental impact and data quality. The score assigned to each node, $Score_n$, is computed as follows:
\begin{equation}
    \small
    Score_n = {W_{E}}\cdot(1 - \frac{CO_{2}^n}{MaxCO_{2}}) + \sum_{i=1}^{I}{W_{i}}\cdot{Q_{i}^n} \label{eq:ranking}
\end{equation}
\noindent where $CO_{2}^n$ represents the environmental footprint of node $n$, while $MaxCO_{2}$ denotes the footprint of the node with the highest environmental impact. Similarly, $Q_{i}^n$ corresponds to the quality dimension $i$ for the dataset in node $n$. The terms $W_{E}$ and $W_{i}$ are weights assigned to the carbon emissions of the node and the data quality properties (e.g., consistency and completeness), respectively. The sum of these weights must be equal to 1. In this study, the weights have been set to prioritize the contribution of energy consumption. Once the scores are computed, nodes can be ranked accordingly.



The predicted number of nodes, $\hat{N}$, must be adapted to fit the FL configuration specified by the researcher. To achieve this, we introduce two key metrics:

\begin{itemize}
    \item \textbf{Basic Data Volume Percentage}, $V_n$: This represents the percentage of the total dataset assigned to a single node $n$ and is defined as:
    \begin{equation}\small
        V_n = \frac{1}{N_C}
    \end{equation}
    where $N_C$ is the total number of nodes in the FL task.
    
    \item \textbf{Target Data Volume Percentage}, $V$: This denotes the total proportion of the dataset recommended for use by the system and is computed as:
    \begin{equation}\small
        V = \hat{N} \cdot V_n
    \end{equation}
\end{itemize}

These metrics ensure that, in a heterogeneous scenario, the predicted number of nodes $\hat{N}$ allows for the selection of the target data volume  necessary $V$ for an effective FL training.

At this stage, the component must select the specific set of nodes to be used in the FL training. Three selection methods are proposed and compared: 
\begin{itemize}
    \item \textbf{Node Selection (NS)}: This method selects the first $\hat{N}$ nodes with the highest score. The selected nodes must satisfy the required data volume percentage $V_n$. If a node $n$ has a dataset larger than the required $V_n$, its data volume is randomly reduced to match $V_n$. Conversely, if a node's data volume does not meet the requirement, it is excluded in favor of the next candidate in the ranking.

    \item \textbf{Minimal Smart Reduction (MSR)}: This method follows the same strategy as NS but incorporates data cleaning by removing low-quality (dirty) data. Only clean data are used for training. As a consequence, the effective data volume $E$, which is the sum of the clean data volumes from the selected nodes, may be lower than the target volume $V$, potentially impacting model accuracy.

    \item \textbf{Smart Reduction (SR)}: This method extends MSR by ensuring that the target data volume $V$ is met. Since removing low-quality data may reduce the total data volume ($E < V$), SR compensates by selecting additional nodes until $E$ reaches or exceeds $V$.
\end{itemize}

The effectiveness of these three methods is evaluated in Sec.~\ref{sec:validation}.

\section{Validation}\label{sec:validation}

The evaluation is conducted by simulating both the role of the researcher and a real-world fog computing scenario. In this context, a set of heterogeneous nodes is considered, each differing in data properties, geographical location, and power consumption. 

The considered scenario sees a researcher with a pre-defined FL configuration aiming at reducing energy costs and optimizing FL training from a data-centric perspective. To support this task, the researcher would use the proposed \textit{FL Configuration Selection System}, providing the inputs listed in Sect.~\ref{subsec:FLRec}, including: the dataset description; the FL participant nodes details; the accuracy estimation; and the accuracy threshold. Additionally, she can provide the weights to be associated with energy and data quality in the scoring of the nodes (Tab.~\ref{tab:thirdinput}).
\begin{table*}[t]
    \small
    \begin{tabularx}{0.65\textwidth}{c c c c c c}
    \toprule
    \textbf{Name} & \textbf{Type} & \textbf{Train Samples} & \textbf{Sequence} & \textbf{Classes} & \textbf{Accuray Estimation} \\
    \midrule
    FreezerRegularTrain & Sensor & 2350 & 301 & 2 & 0.81 \\
    TwoLeadECG & ECG & 1039 & 82 & 2 & 0.70 \\
    ElectricDevices & Device & 8894 & 96 & 7 & 0.53 \\
    \bottomrule
    \end{tabularx}
    \caption{Evaluation Datasets.}
    \label{tab:evaluationdataset}
\end{table*}
\begin{table}[t]
    \centering
    \begin{tabular}{c c c}
    \toprule
    \textbf{Energy} & \textbf{Consistency} & \textbf{Completeness}\\
    \midrule
    0.7 & 0.2 & 0.1 \\
    \bottomrule
    \end{tabular}
    \\
    \caption{Score Weights Example}
    \label{tab:thirdinput}
\end{table}

For evaluation purposes, three distinct FL initial configurations have been tested using three different datasets (Tab.~\ref{tab:evaluationdataset}). These configurations, referred to as Configuration 1, Configuration 2, and Configuration 3 (Tab.~\ref{tab:combinedconfig}), incorporate heterogeneous nodes distributed across different global regions, each with varying hardware capabilities and data quality levels. The power-related measurements are derived from the real specifications of existing GPU and TPU devices. While the carbon intensity for each location is not explicitly listed, it is obtained from \cite{electricitymap} and utilized to compute the actual carbon emissions generated during training. Each FL configuration is paired with a specific dataset: FreezerRegularTrain, TwoLeadECG, and ElectricDevices, respectively.

The initial accuracy estimation is computed by selecting one random client and training it for a full FL cycle with just one node. The accuracy estimation performance value reports only an indication of the pattern and capabilities of satisfying the task, therefore only an approximated accuracy value is returned. 

\begin{table*}[t]
    \small
    \centering
    \begin{tabularx}{0.75\textwidth}{l | c c c c c}
    \toprule
    \textbf{Configuration ID} & \textbf{Power (kWh)} & \textbf{Location} & \textbf{Data Volume} & \textbf{Consistency} & \textbf{Completeness} \\
    \midrule
    \textbf{Configuration 1} & 350 & Finland & 0.11 & 0.90 & 0.90 \\
    & 10 & Germany & 0.07 & 0.90 & 0.90 \\
    & 75 & Portugal & 0.064 & 0.95 & 0.80 \\
    & 250 & Portugal & 0.08 & 0.70 & 0.70 \\
    & 100 & Canada & 0.12 & 0.95 & 0.50 \\
    & 350 & California & 0.075 & 0.95 & 0.50 \\
    & 300 & Bosnia Herzegovina & 0.09 & 0.60 & 0.90 \\
    & 75 & Finland & 0.08 & 0.95 & 0.95 \\
    & 30 & California & 0.068 & 0.80 & 0.80 \\
    & 10 & Bosnia Herzegovina & 0.094 & 0.90 & 0.85 \\
    & 100 & Germany & 0.088 & 0.85 & 0.93 \\
    & 250 & Finland & 0.061 & 0.93 & 0.91 \\
    \midrule
    \textbf{Configuration 2} & 350 & Finland & 0.18 & 0.94 & 0.90 \\
    & 10 & Germany & 0.10 & 0.80 & 0.95 \\
    & 75 & Portugal & 0.20 & 0.95 & 0.87 \\
    & 250 & Portugal & 0.12 & 0.70 & 0.85 \\
    & 100 & Canada & 0.10 & 0.95 & 0.80 \\
    & 300 & Bosnia Herzegovina & 0.15 & 0.60 & 0.90 \\
    & 75 & Finland & 0.09 & 0.97 & 0.75 \\
    & 10 & Bosnia Herzegovina & 0.06 & 0.90 & 0.80 \\
    \midrule
    \textbf{Configuration 3} & 350 & Finland & 0.18 & 0.78 & 0.80 \\
    & 10 & Germany & 0.10 & 0.87 & 0.89 \\
    & 75 & Portugal & 0.20 & 0.93 & 0.95 \\
    & 250 & Portugal & 0.12 & 0.70 & 0.70 \\
    & 100 & Canada & 0.10 & 0.91 & 0.80 \\
    & 350 & California & 0.10 & 0.95 & 0.70 \\
    & 300 & Bosnia Herzegovina & 0.05 & 0.82 & 0.82 \\
    & 75 & Finland & 0.05 & 0.95 & 0.95 \\
    & 30 & California & 0.04 & 0.90 & 0.90 \\
    & 10 & Bosnia Herzegovina & 0.06 & 0.98 & 0.98 \\
    \bottomrule
    \end{tabularx}
    \caption{FL Configuration Data for Three Scenarios.}
    \label{tab:combinedconfig}
\end{table*}

Using the described datasets and configurations, the three proposed approaches have been tested and compared against a baseline. The \textbf{Baseline} approach is defined as training without any optimization techniques applied; thus, it corresponds to training the original FL configuration using all available resources and data. The results present a direct comparison of the four methods: \textbf{Baseline}, \textbf{NS}, \textbf{SR}, and \textbf{MSR}. The evaluation is based on two key metrics relevant to this study: (i) the carbon emissions produced by the overall FL system during both the pre-processing and training phases, and (ii) the final accuracy achieved in predicting a disjoint test set by the trained FL configuration. The results will be analyzed by evaluating each configuration individually, comparing final accuracy and carbon emissions. The validation experiments have been conducted by running and repeating each experiment 8 times. Given that there are three FL configurations, three proposed methods, and the Baseline approach applied to each configuration, the total number of simulations performed for the validation amounts to 96.

\subsection{Configuration 1}
\label{subsec:conf1}
The first evaluation experiment was conducted using Configuration 1 (Tab.~\ref{tab:combinedconfig}) with the FreezerRegularTrain dataset (Tab.~\ref{tab:evaluationdataset}). As a first step, the accuracy estimation was performed, yielding a result of 0.81 with emissions of 0.03 kg~$CO_{2}$e. As a realistic objective, the accuracy threshold set by the researcher was assumed to be 0.85.

\begin{figure}[t]
    \centering
    \includegraphics[width=0.98\columnwidth]{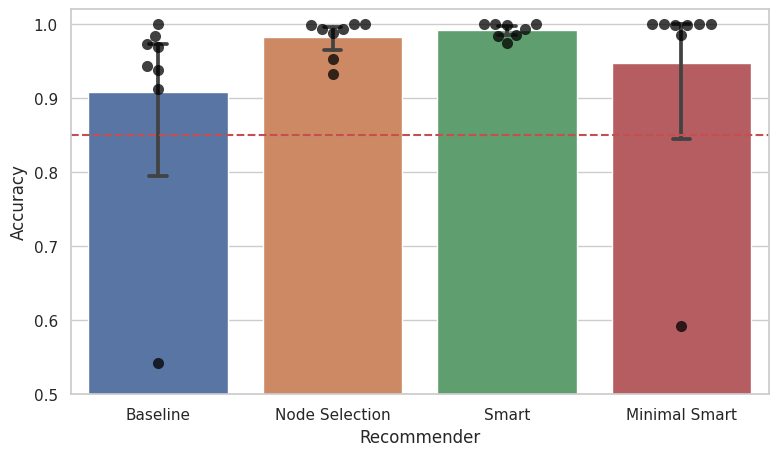}
    \caption{Configuration 1 Accuracy Results.}
    \label{fig:conf1_accuracy}
\end{figure}

\begin{figure}[t]
    \centering
    \includegraphics[width=0.98\columnwidth]{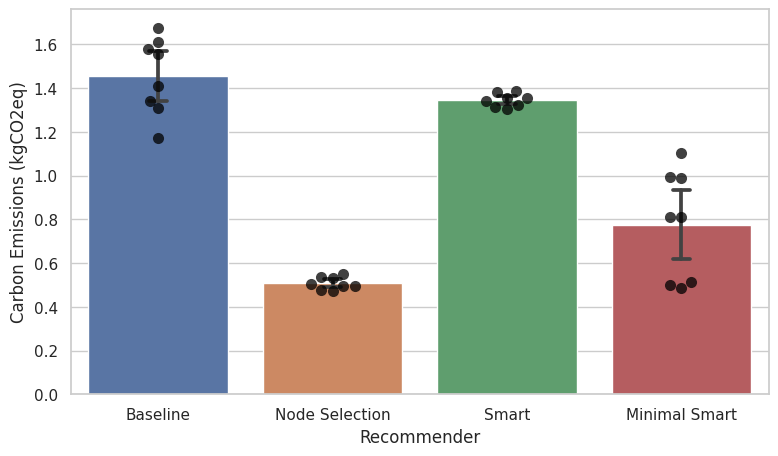}
    \caption{Configuration 1 Carbon Emissions Results.}
    \label{fig:conf1_energy}
\end{figure}

The results are presented in Fig.~\ref{fig:conf1_accuracy}, which illustrates the distribution of accuracies obtained using the Baseline and the three proposed recommender systems. The red dashed line represents the accuracy threshold set at 0.85. The plot demonstrates that the Baseline approach, which utilizes all available resources (i.e., all nodes without dataset manipulation), meets the accuracy constraint seven times out of eight. Notably, the three recommender systems achieve even better results while using fewer resources. The Minimal Smart Reduction recommender, in particular, satisfies the researcher's constraint in seven out of eight cases, matching the Baseline. Overall, a slight improvement in accuracy can be observed with a reduced resource footprint, particularly when leveraging cleaner data.

The comparison of carbon emissions impact is shown in Fig.~\ref{fig:conf1_energy}. As expected, the Baseline, utilizing all nodes and the entire dataset, is the most energy-intensive method, reaching over 1.4 kg~$CO_{2}$e. The NS method, which selects only 6 out of 10 clients without data removal, proves to be more efficient, reducing emissions by approximately 1.0 kg~$CO_{2}$e. The SR method involves fewer nodes than the Baseline; however, its emissions remain substantial due to the number of nodes engaged. Specifically, in Configuration 1, SR selects 10 clients, similar to the Baseline, but applies data removal. To meet the required total data volume of 0.6, additional client nodes must participate in the FL training. The MSR method employs the same set and number of nodes as Node Selection but trains exclusively on clean data. This difference impacts the number of epochs required and increases the overall training duration, leading to higher carbon emissions. This behavior is attributed to the early stopping technique: training on clean data results in a consistent reduction of evaluation loss after each round, preventing premature termination of the training process. In contrast, training with dirty data causes fluctuations in loss, triggering early stopping and reducing the number of epochs executed.

In Configuration 1, all methods successfully met the accuracy threshold, except for a single experiment using the MSR approach. Additionally, every experiment demonstrated a reduction in carbon emissions compared to the Baseline. Among the proposed methods, the NS algorithm appears to be the most effective, achieving the best balance between accuracy and energy efficiency. However, the MSR method exhibits a slight improvement in accuracy while maintaining a comparable level of carbon emissions.

\subsection{Configuration 2}
\label{subsec:conf2}

The second evaluation experiment was conducted using Configuration 2 from Tab.~\ref{tab:combinedconfig}, paired with the TwoLeadECG dataset from Tab.~\ref{tab:evaluationdataset}. The measured accuracy estimation resulted in 0.70, while the carbon emissions generated during computation amounted to 0.071 kg$CO_{2}$e. As in the previous experiment, the desired accuracy threshold was set to 0.85.

\begin{figure}[t]
    \centering
    \includegraphics[width=0.98\columnwidth]{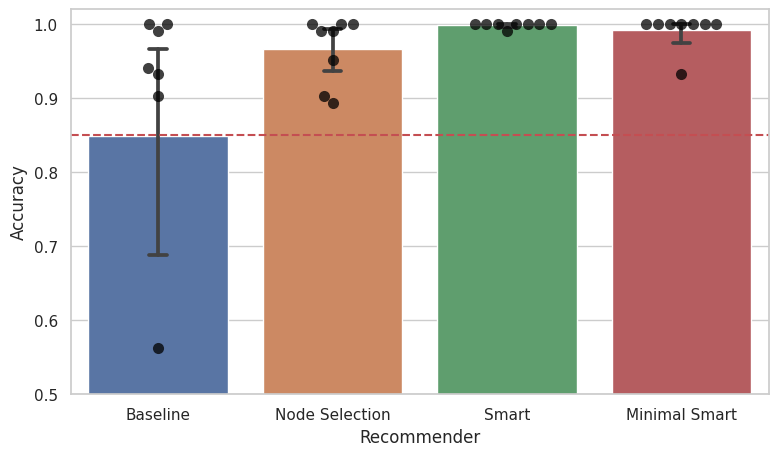}
    \caption{Configuration 2 Accuracy Results.}
    \label{fig:conf2_accuracy}
\end{figure}

\begin{figure}[t]
    \centering
    \includegraphics[width=0.98\columnwidth]{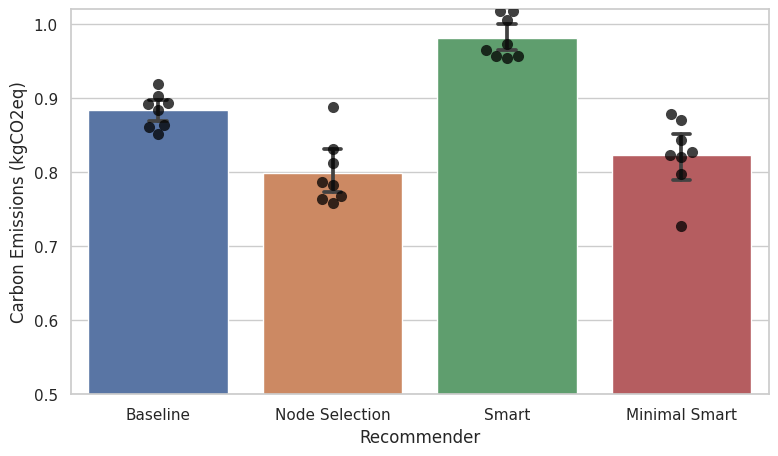}
    \caption{Configuration 2 Carbon Emissions Results.}
    \label{fig:conf2_energy}
\end{figure}

The pattern observed in Fig.~\ref{fig:conf2_accuracy} closely resembles the one described in Sec.~\ref{subsec:conf1} regarding accuracy results. The Baseline exhibits the worst performance among all approaches, failing to meet the required accuracy in one of the experiments. The reduction in resource usage, particularly through data removal, proves beneficial and leads to improved performance in this FL environment. Notably, the NS and MSR methods selected 5 clients, whereas the SR method required all 8 clients defined in Configuration 2.

In Configuration 2, specific characteristics of the dataset must be highlighted. The TwoLeadECG dataset used in this experiment contains only 1,039 samples with a sequence length of 82, as outlined in Tab.~\ref{tab:evaluationdataset}. Given that these samples are distributed across 8 nodes, the overall dataset size per node remains low. Consequently, the reduction in carbon emissions achieved by the proposed methods is limited. Compared to the Baseline in Sec.~\ref{subsec:conf1}, the Baseline in this experiment produces lower carbon emissions. However, the $CO_{2}$ emissions of the SR method exceed those of the Baseline due to the selection of all available nodes to meet the target data volume requirement introduced in Sec.~\ref{subsec:recommender}. In fact, in this scenario, data removal and the use of clean data prevent early stopping, resulting in a higher number of training epochs, as previously observed in the validation of Configuration 1. The MSR algorithm delivers results similar to the NS method, selecting identical nodes and achieving an average reduction in emissions of 10\%. 

In Configuration 2, all recommended configurations successfully met the accuracy threshold. However, not all of them reduced the carbon footprint compared to the Baseline, as the SR method failed in this regard. The MSR approach demonstrated notable improvements in accuracy while maintaining a carbon emissions level comparable to the NS method.

\subsection{Configuration 3}

\begin{figure}[!t]
    \centering
    \includegraphics[width=0.98\columnwidth]{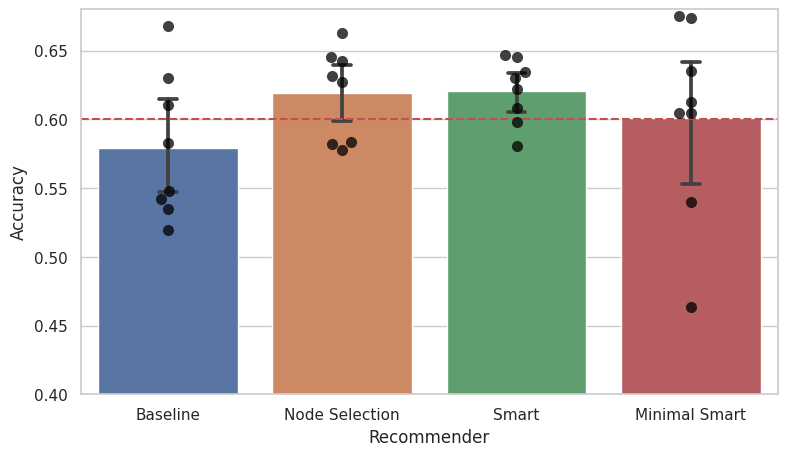}
    \caption{Configuration 3 Accuracy Results.}
    \label{fig:conf3_accuracy}
\end{figure}

\begin{figure}[!t]
    \centering
    \includegraphics[width=0.98\columnwidth]{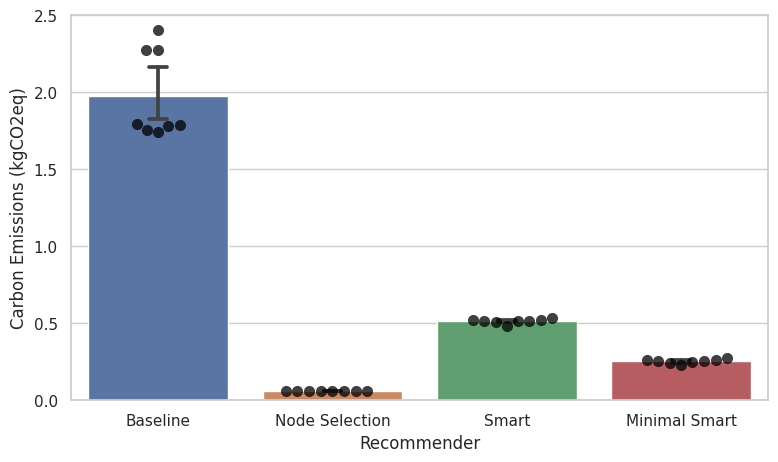}
    \caption{Configuration 3 Energy Results.}
    \label{fig:conf3_energy}
\end{figure}

In Configuration 3, the task proves to be more complex than in Configurations 1 and 2, likely due to the larger number of training samples and the increased number of classes. This complexity is evident when observing the maximum accuracy value, which does not exceed 0.66, as shown in Fig.~\ref{fig:conf3_accuracy}. Additionally, Configuration 3 results in the highest carbon emissions among all observed configurations, as illustrated in Fig.~\ref{fig:conf3_energy}. The accuracy estimation yields a value of 0.54, while the carbon emissions generated for the estimation amount to 0.025 kg$CO_{2}$e. For this experiment, the accuracy threshold is set to 0.60, a target that exceeds the accuracy achieved in five of the experiments conducted using the Baseline approach.

The pattern shown in Fig.~\ref{fig:conf3_accuracy} is similar to the ones already described. The Baseline shows the worst performance, failing to satisfy the required accuracy 5 out of 8 times. The usage of fewer resources, especially applying data removal is beneficial and leads to an improvement in performance in this FL environment. The NS and the MSR methods have selected only 2 nodes, while the SR method selected 3 clients to satisfy the data volume requirement. All the methods show improvements in accuracy compared to the Baseline, while still failing to reach the required accuracy in some experiments (NS reaches the goal 5 out of 8 times, while SR and MSR succeed 6 out of 8 times).

In Configuration 3, the Baseline carbon emissions is relevant, reaching 2.5 kg$CO_{2}$e. The dataset used in Configuration 3 has more than 8000 samples to train, resulting in the most demanding FL training compared with the previous configurations. The recommender systems significantly reduce of 80$\%$ the carbon emissions generation on average. The NS method is the most effective, reducing the number of nodes without applying removal of low quality data. The MSR produces slightly larger emissions with respect to NS, even though it has a higher energy cost probably due to dirty data removal. Training with less data should theoretically yield an advantage in terms of emissions. However, data removal, while eliminating dirty data, leads to training with a reduced but cleaner dataset. This, in turn, helps avoid early stopping and allows for the execution of more epochs, ultimately increasing the duration time and energy consumption.

In Configuration 3, the recommender systems presented have satisfied the accuracy threshold with 70$\%$ of success, improving on average the accuracy returned by the Baseline method. All of them have significantly reduced the carbon impact compared to the Baseline. The MSR method has obtained slightly worst results compared with NS in carbon emissions generation, but it has reached better accuracy levels.

\subsection{Final Analysis}

After comparing the three developed methods with the Baseline across the proposed configurations, general conclusions and observations can now be drawn.

\paragraph{Accuracy Results}
All three methods performed well in all scenarios, showing an improvement over the Baseline. The NS method demonstrates slightly better results than the Baseline due to its prioritization of nodes with a volume satisfying the Basic Data Volume Percentage. This strategy retains the original (potentially noisy) data. The NS method achieves a 10\% improvement in accuracy, meeting the accuracy threshold set by the researcher in 87\% of the experiments. The MSR method selects the same nodes as NS but applies data filtering, using only clean data during FL training. This does not drastically alter the training performance compared to NS. In general, MSR registers an 8\% accuracy improvement over the Baseline, satisfying the accuracy threshold in 87\% of cases. The SR method, which selects more nodes with clean data than the previous approaches, yields the best accuracy results. It reports a 12\% improvement over the Baseline and meets the accuracy threshold in 92\% of the experiments. From an accuracy perspective, SR is the most effective method. From these observations, we can conclude that the desired accuracy can be achieved and even improved while using a lower volume of data. The proposed recommender system proved effective in suggesting the appropriate data volume required to reach the target accuracy level. Furthermore, removing low-quality data enhances accuracy, but only if the target data volume is maintained.

\paragraph{Carbon Emissions}
NS achieves the best energy efficiency, reducing carbon emissions by an average of 56\% compared to the Baseline, with a peak reduction of approximately 90\%. MSR, which selects the same nodes as NS but uses clean data, results in longer FL training due to an increased number of epochs and extended duration. Consequently, its average reduction in emissions is 45\%, which is lower than that of NS, though it consistently produces lower $CO_2$ emissions than the Baseline.
The SR method, which selects more nodes (sometimes the same number as the Baseline), has a lesser impact on emissions, achieving an average reduction of only 25\%. Thus, NS outperforms the other methods in terms of energy efficiency.

\paragraph{Overall Analysis}
As a final analysis, both NS and MSR improve accuracy over the Baseline while consistently minimizing energy consumption, with NS performing best in carbon footprint reduction. Meanwhile, SR excels in improving FL model accuracy but with a lower improvement in the environmental impact. In conclusion, after reviewing the validation results, it can be stated that the NS method provides the best FL configuration recommendation.

\paragraph{Life Cycle Emissions}

As detailed in Sec.~\ref{subsec:exploration_impl}, extensive experiments were conducted to extract insights for developing the \textit{FL Reduction System}. The \textit{FL Simulator} accounted for the highest carbon emissions, as all training sessions were executed within it.

The first set of experiments analyzed data volume reduction using five datasets with horizontal and vertical reduction methods. Each reduction type was tested across five configurations (100\%, 80\%, ..., 20\%) and repeated three times to mitigate randomness, resulting in 150 FL training simulations and 1.93 kg$CO_2$e emissions. The second set focused on data quality, evaluating three quality dimensions across both reduction approaches using a single dataset. These experiments were also repeated three times, totaling 90 simulations and 0.54 kg$CO_2$e emissions.

Overall, the \textit{Data-Centric FL Exploration} phase generated 2.47 kg$CO_2$e. While this process is energy-intensive, it does not need to be repeated for every training task. Instead, the insights and \textit{FL Reduction System} developed can be reused by multiple researchers, making the methodology sustainable in the medium to long term.

\subsection{Transparency and Reproducibility}

To ensure transparency and reproducibility, we provide a practical implementation of our proposed solution, including the approach and architecture, available at the following repository: \url{https://github.com/POLIMIGreenISE/ecoFL.git}. This repository contains all necessary resources to replicate our experiments and validate our findings.

The implementation is entirely developed in Python~\cite{python}, leveraging well-established libraries such as NumPy~\cite{numpy} and Pandas~\cite{pandas} for data manipulation, including arrays, data frames, and multidimensional matrices. Scikit-Learn~\cite{scikit} has been employed for creating baseline machine learning models, such as linear regression. Data visualization is facilitated using Matplotlib~\cite{matplotlib} and Streamlit~\cite{streamlit}, which also serves as the framework for developing the solution as a service. This approach enables seamless and immediate execution of simulations, as well as inference and validation processes.

Our \textit{FL Simulator} is built using Flower~\cite{flower}, a federated learning framework that enables the simulation of multiple nodes on a single system. Flower provides essential functionalities for parameter transmission and aggregation during training. The deep learning model used in our experiments is a ResNet architecture, implemented using the Keras TensorFlow library.

The primary experimental task is Time Series Classification and several datasets from the UCR/UEA archive~\cite{uea, ucr}, have been used during the Data-Centric FL Exploration and Validation phase.

Comprehensive documentation and detailed execution guidelines are provided in the repository's \texttt{README.md} file. By making all code, data, and experimental procedures publicly accessible, we aim to enhance the reproducibility of our findings and support further research in sustainable Federated Learning.

\section{Conclusion}\label{sec:conclusion}

This work explored a data-centric approach to energy-efficient Federated Learning (FL), focusing on optimizing data volume and node selection for heterogeneous and distributed datasets. By analyzing the relationship between data-centric attributes, energy consumption, and FL model performance, we developed an approach that effectively minimizes carbon emissions while maintaining predictive accuracy.  

We introduced the \textit{FL Reduction System}, designed to support researchers in selecting optimal FL configurations that reduce environmental impact without compromising model performance. Through extensive experimentation across three FL configurations, we demonstrated that efficient data management strategies can significantly reduce computational costs. Among the proposed methods, \textbf{Node Selection} proved the most effective in balancing accuracy and energy efficiency, achieving an average 56\% reduction in carbon emissions. \textbf{Minimal Smart Reduction} further incorporated data quality filtering, while \textbf{Smart Reduction} prioritized accuracy gains but was less efficient in reducing emissions. These findings highlight the critical role of federated and cloud-based data management in optimizing distributed ML workflows.  

Beyond individual FL training tasks, this study contributes to \textbf{data management support for ML} by providing a reusable knowledge base for future FL deployments. The insights gained can inform adaptive strategies for handling heterogeneous datasets, optimizing cloud data management, and reducing the environmental footprint of FL training.  

Future research could refine the proposed approach by dynamically adjusting ranking weights for balancing energy efficiency and data quality, rather than using fixed values. Additionally, exploring the impact of FL hyperparameters—such as batch size, early stopping, number of epochs, and aggregation techniques—could lead to further efficiency gains. Investigating data distribution effects, both system-wide and within individual nodes, from an energy perspective would also enhance sustainability. Finally, incorporating additional data quality dimensions could further improve FL performance while reducing data volume, strengthening the impact of energy-efficient data systems in federated and cloud-based learning environments.

\begin{acks}
This work was supported by the project FREEDA, funded by the frameworks PRIN (MUR, Italy) and by the European Union (TEADAL, 101070186).
\end{acks}


\bibliographystyle{ACM-Reference-Format}
\bibliography{ref.bib}


\begin{thebibliography}{53}


\ifx \showCODEN    \undefined \def \showCODEN     #1{\unskip}     \fi
\ifx \showDOI      \undefined \def \showDOI       #1{#1}\fi
\ifx \showISBNx    \undefined \def \showISBNx     #1{\unskip}     \fi
\ifx \showISBNxiii \undefined \def \showISBNxiii  #1{\unskip}     \fi
\ifx \showISSN     \undefined \def \showISSN      #1{\unskip}     \fi
\ifx \showLCCN     \undefined \def \showLCCN      #1{\unskip}     \fi
\ifx \shownote     \undefined \def \shownote      #1{#1}          \fi
\ifx \showarticletitle \undefined \def \showarticletitle #1{#1}   \fi
\ifx \showURL      \undefined \def \showURL       {\relax}        \fi
\providecommand\bibfield[2]{#2}
\providecommand\bibinfo[2]{#2}
\providecommand\natexlab[1]{#1}
\providecommand\showeprint[2][]{arXiv:#2}

\bibitem[\protect\citeauthoryear{??}{ele}{ly 1}]%
        {electricitymap}
 \bibinfo{year}{2023, July 1}\natexlab{}.
\newblock \bibinfo{booktitle}{\emph{Electricity Maps - Data Portal.}}
\newblock \bibinfo{type}{{T}echnical {R}eport}.
\newblock
\urldef\tempurl%
\url{https://www.electricitymaps.com/data-portal}
\showURL{%
\tempurl}


\bibitem[\protect\citeauthoryear{Abbasi, Dong, Wang, Leung, Zhou, and Drew}{Abbasi et~al\mbox{.}}{2024}]%
        {abbasi2024fedgreencarbonawarefederatedlearning}
\bibfield{author}{\bibinfo{person}{Ali Abbasi}, \bibinfo{person}{Fan Dong}, \bibinfo{person}{Xin Wang}, \bibinfo{person}{Henry Leung}, \bibinfo{person}{Jiayu Zhou}, {and} \bibinfo{person}{Steve Drew}.} \bibinfo{year}{2024}\natexlab{}.
\newblock \showarticletitle{FedGreen: Carbon-aware Federated Learning with Model Size Adaptation}. In \bibinfo{booktitle}{\emph{2024 IEEE International Conference on Communications Workshops (ICC Workshops)}}. IEEE, \bibinfo{pages}{1352--1358}.
\newblock


\bibitem[\protect\citeauthoryear{Anselmo and Vitali}{Anselmo and Vitali}{2023}]%
        {anselmo2023data}
\bibfield{author}{\bibinfo{person}{Mart{\'\i}n Anselmo} {and} \bibinfo{person}{Monica Vitali}.} \bibinfo{year}{2023}\natexlab{}.
\newblock \showarticletitle{A data-centric approach for reducing carbon emissions in deep learning}. In \bibinfo{booktitle}{\emph{International Conference on Advanced Information Systems Engineering}}. Springer, \bibinfo{pages}{123--138}.
\newblock


\bibitem[\protect\citeauthoryear{Bagnall, Dau, Lines, Flynn, Large, Bostrom, Southam, and Keogh}{Bagnall et~al\mbox{.}}{2018}]%
        {uea}
\bibfield{author}{\bibinfo{person}{Anthony Bagnall}, \bibinfo{person}{Hoang~Anh Dau}, \bibinfo{person}{Jason Lines}, \bibinfo{person}{Michael Flynn}, \bibinfo{person}{James Large}, \bibinfo{person}{Aaron Bostrom}, \bibinfo{person}{Paul Southam}, {and} \bibinfo{person}{Eamonn Keogh}.} \bibinfo{year}{2018}\natexlab{}.
\newblock \bibinfo{title}{The UEA multivariate time series classification archive, 2018}.
\newblock
\newblock
\showeprint[arxiv]{1811.00075}~[cs.LG]


\bibitem[\protect\citeauthoryear{Berti-Equille}{Berti-Equille}{2019}]%
        {berti2019learn2clean}
\bibfield{author}{\bibinfo{person}{Laure Berti-Equille}.} \bibinfo{year}{2019}\natexlab{}.
\newblock \showarticletitle{{Learn2clean: Optimizing the Sequence of Tasks for Web Data Preparation}}. In \bibinfo{booktitle}{\emph{The World Wide Web Conference}}. \bibinfo{pages}{2580--2586}.
\newblock


\bibitem[\protect\citeauthoryear{Beutel, Topal, Mathur, Qiu, Fernandez-Marques, Gao, Sani, Kwing, Parcollet, Gusmão, and Lane}{Beutel et~al\mbox{.}}{2020}]%
        {flower}
\bibfield{author}{\bibinfo{person}{Daniel~J Beutel}, \bibinfo{person}{Taner Topal}, \bibinfo{person}{Akhil Mathur}, \bibinfo{person}{Xinchi Qiu}, \bibinfo{person}{Javier Fernandez-Marques}, \bibinfo{person}{Yan Gao}, \bibinfo{person}{Lorenzo Sani}, \bibinfo{person}{Hei~Li Kwing}, \bibinfo{person}{Titouan Parcollet}, \bibinfo{person}{Pedro PB~de Gusmão}, {and} \bibinfo{person}{Nicholas~D Lane}.} \bibinfo{year}{2020}\natexlab{}.
\newblock \showarticletitle{Flower: A Friendly Federated Learning Research Framework}.
\newblock \bibinfo{journal}{\emph{arXiv preprint arXiv:2007.14390}} (\bibinfo{year}{2020}).
\newblock


\bibitem[\protect\citeauthoryear{Bonawitz, Eichner, Grieskamp, Huba, Ingerman, Ivanov, Kiddon, Kone{\v{c}}n{\`y}, Mazzocchi, McMahan, et~al\mbox{.}}{Bonawitz et~al\mbox{.}}{2019}]%
        {bonawitz2019federatedlearningscaledesign}
\bibfield{author}{\bibinfo{person}{Keith Bonawitz}, \bibinfo{person}{Hubert Eichner}, \bibinfo{person}{Wolfgang Grieskamp}, \bibinfo{person}{Dzmitry Huba}, \bibinfo{person}{Alex Ingerman}, \bibinfo{person}{Vladimir Ivanov}, \bibinfo{person}{Chloe Kiddon}, \bibinfo{person}{Jakub Kone{\v{c}}n{\`y}}, \bibinfo{person}{Stefano Mazzocchi}, \bibinfo{person}{Brendan McMahan}, {et~al\mbox{.}}} \bibinfo{year}{2019}\natexlab{}.
\newblock \showarticletitle{Towards federated learning at scale: System design}.
\newblock \bibinfo{journal}{\emph{Proceedings of machine learning and systems}}  \bibinfo{volume}{1} (\bibinfo{year}{2019}), \bibinfo{pages}{374--388}.
\newblock


\bibitem[\protect\citeauthoryear{Budach, Feuerpfeil, Ihde, Nathansen, Noack, Patzlaff, Naumann, and Harmouch}{Budach et~al\mbox{.}}{2022}]%
        {budach2022effects}
\bibfield{author}{\bibinfo{person}{Lukas Budach}, \bibinfo{person}{Moritz Feuerpfeil}, \bibinfo{person}{Nina Ihde}, \bibinfo{person}{Andrea Nathansen}, \bibinfo{person}{Nele Noack}, \bibinfo{person}{Hendrik Patzlaff}, \bibinfo{person}{Felix Naumann}, {and} \bibinfo{person}{Hazar Harmouch}.} \bibinfo{year}{2022}\natexlab{}.
\newblock \showarticletitle{The effects of data quality on machine learning performance}.
\newblock \bibinfo{journal}{\emph{arXiv preprint arXiv:2207.14529}} (\bibinfo{year}{2022}).
\newblock


\bibitem[\protect\citeauthoryear{Castanyer, Mart{\'\i}nez-Fern{\'a}ndez, and Franch}{Castanyer et~al\mbox{.}}{2024}]%
        {castanyer2021design}
\bibfield{author}{\bibinfo{person}{Roger~Creus Castanyer}, \bibinfo{person}{Silverio Mart{\'\i}nez-Fern{\'a}ndez}, {and} \bibinfo{person}{Xavier Franch}.} \bibinfo{year}{2024}\natexlab{}.
\newblock \showarticletitle{Which design decisions in AI-enabled mobile applications contribute to greener AI?}
\newblock \bibinfo{journal}{\emph{Empirical Software Engineering}} \bibinfo{volume}{29}, \bibinfo{number}{1} (\bibinfo{year}{2024}), \bibinfo{pages}{2}.
\newblock


\bibitem[\protect\citeauthoryear{Chai, Jin, Tang, Fan, Miao, Wang, Luo, Li, Yuan, and Wang}{Chai et~al\mbox{.}}{2025}]%
        {chai2025cost}
\bibfield{author}{\bibinfo{person}{Chengliang Chai}, \bibinfo{person}{Kaisen Jin}, \bibinfo{person}{Nan Tang}, \bibinfo{person}{Ju Fan}, \bibinfo{person}{Dongjing Miao}, \bibinfo{person}{Jiayi Wang}, \bibinfo{person}{Yuyu Luo}, \bibinfo{person}{Guoliang Li}, \bibinfo{person}{Ye Yuan}, {and} \bibinfo{person}{Guoren Wang}.} \bibinfo{year}{2025}\natexlab{}.
\newblock \showarticletitle{Cost-effective Missing Value Imputation for Data-effective Machine Learning}.
\newblock \bibinfo{journal}{\emph{ACM Transactions on Database Systems}} (\bibinfo{year}{2025}).
\newblock


\bibitem[\protect\citeauthoryear{Chai, Wang, Luo, Niu, and Li}{Chai et~al\mbox{.}}{2023}]%
        {9705125}
\bibfield{author}{\bibinfo{person}{Chengliang Chai}, \bibinfo{person}{Jiayi Wang}, \bibinfo{person}{Yuyu Luo}, \bibinfo{person}{Zeping Niu}, {and} \bibinfo{person}{Guoliang Li}.} \bibinfo{year}{2023}\natexlab{}.
\newblock \showarticletitle{Data Management for Machine Learning: A Survey}.
\newblock \bibinfo{journal}{\emph{IEEE Transactions on Knowledge and Data Engineering}} \bibinfo{volume}{35}, \bibinfo{number}{5} (\bibinfo{year}{2023}), \bibinfo{pages}{4646--4667}.
\newblock
\urldef\tempurl%
\url{https://doi.org/10.1109/TKDE.2022.3148237}
\showDOI{\tempurl}


\bibitem[\protect\citeauthoryear{Cormode, Markov, and Srinivas}{Cormode et~al\mbox{.}}{2024}]%
        {cormode2024private}
\bibfield{author}{\bibinfo{person}{Graham Cormode}, \bibinfo{person}{Igor~L Markov}, {and} \bibinfo{person}{Harish Srinivas}.} \bibinfo{year}{2024}\natexlab{}.
\newblock \showarticletitle{Private and Efficient Federated Numerical Aggregation.}. In \bibinfo{booktitle}{\emph{EDBT}}. \bibinfo{pages}{734--742}.
\newblock


\bibitem[\protect\citeauthoryear{Frey, Zhao, Axelrod, Jones, Bestor, Gadepally, G{\'o}mez-Bombarelli, and Samsi}{Frey et~al\mbox{.}}{2022}]%
        {frey2022energy}
\bibfield{author}{\bibinfo{person}{Nathan~C Frey}, \bibinfo{person}{Dan Zhao}, \bibinfo{person}{Simon Axelrod}, \bibinfo{person}{Michael Jones}, \bibinfo{person}{David Bestor}, \bibinfo{person}{Vijay Gadepally}, \bibinfo{person}{Rafael G{\'o}mez-Bombarelli}, {and} \bibinfo{person}{Siddharth Samsi}.} \bibinfo{year}{2022}\natexlab{}.
\newblock \showarticletitle{{Energy-aware Neural Architecture Selection and Hyperparameter Optimization}}. In \bibinfo{booktitle}{\emph{2022 IEEE International Parallel and Distributed Processing Symposium Workshops (IPDPSW)}}. IEEE, \bibinfo{pages}{732--741}.
\newblock


\bibitem[\protect\citeauthoryear{Georgiou et~al\mbox{.}}{Georgiou et~al\mbox{.}}{2022}]%
        {georgiou2022green}
\bibfield{author}{\bibinfo{person}{Stefanos Georgiou} {et~al\mbox{.}}} \bibinfo{year}{2022}\natexlab{}.
\newblock \showarticletitle{Green ai: Do deep learning frameworks have different costs?}. In \bibinfo{booktitle}{\emph{Proceedings of the 44th International Conference on Software Engineering}}. \bibinfo{pages}{1082--1094}.
\newblock


\bibitem[\protect\citeauthoryear{Gupta, Mujumdar, Patel, Masuda, Panwar, Bandyopadhyay, Mehta, Guttula, Afzal, Sharma~Mittal, et~al\mbox{.}}{Gupta et~al\mbox{.}}{2021}]%
        {gupta2021data}
\bibfield{author}{\bibinfo{person}{Nitin Gupta}, \bibinfo{person}{Shashank Mujumdar}, \bibinfo{person}{Hima Patel}, \bibinfo{person}{Satoshi Masuda}, \bibinfo{person}{Naveen Panwar}, \bibinfo{person}{Sambaran Bandyopadhyay}, \bibinfo{person}{Sameep Mehta}, \bibinfo{person}{Shanmukha Guttula}, \bibinfo{person}{Shazia Afzal}, \bibinfo{person}{Ruhi Sharma~Mittal}, {et~al\mbox{.}}} \bibinfo{year}{2021}\natexlab{}.
\newblock \showarticletitle{Data quality for machine learning tasks}. In \bibinfo{booktitle}{\emph{Proceedings of the 27th ACM SIGKDD conference on knowledge discovery \& data mining}}. \bibinfo{pages}{4040--4041}.
\newblock


\bibitem[\protect\citeauthoryear{Harris, Millman, van~der Walt, et~al\mbox{.}}{Harris et~al\mbox{.}}{2020}]%
        {numpy}
\bibfield{author}{\bibinfo{person}{Charles~R. Harris}, \bibinfo{person}{K.~Jarrod Millman}, \bibinfo{person}{St{\'{e}}fan~J. van~der Walt}, {et~al\mbox{.}}} \bibinfo{year}{2020}\natexlab{}.
\newblock \showarticletitle{Array programming with {NumPy}}.
\newblock \bibinfo{journal}{\emph{Nature}} \bibinfo{volume}{585}, \bibinfo{number}{7825} (\bibinfo{date}{sep} \bibinfo{year}{2020}), \bibinfo{pages}{357--362}.
\newblock
\urldef\tempurl%
\url{https://doi.org/10.1038/s41586-020-2649-2}
\showDOI{\tempurl}


\bibitem[\protect\citeauthoryear{He et~al\mbox{.}}{He et~al\mbox{.}}{2016}]%
        {resnet}
\bibfield{author}{\bibinfo{person}{Kaiming He} {et~al\mbox{.}}} \bibinfo{year}{2016}\natexlab{}.
\newblock \showarticletitle{{Deep Residual Learning for Image Recognition}}. In \bibinfo{booktitle}{\emph{Proceedings of the IEEE Conference on Computer Vision and Pattern Recognition}}. \bibinfo{pages}{770--778}.
\newblock


\bibitem[\protect\citeauthoryear{Hoang, Lekssays, Carminati, Ferrari, et~al\mbox{.}}{Hoang et~al\mbox{.}}{2023}]%
        {hoang2023privacy}
\bibfield{author}{\bibinfo{person}{Anh-Tu Hoang}, \bibinfo{person}{Ahmed Lekssays}, \bibinfo{person}{Barbara Carminati}, \bibinfo{person}{Elena Ferrari}, {et~al\mbox{.}}} \bibinfo{year}{2023}\natexlab{}.
\newblock \showarticletitle{Privacy-preserving Decentralized Learning of Knowledge Graph Embeddings.}. In \bibinfo{booktitle}{\emph{EDBT/ICDT Workshops}}.
\newblock


\bibitem[\protect\citeauthoryear{Hsiao et~al\mbox{.}}{Hsiao et~al\mbox{.}}{2019}]%
        {Hsiao2019}
\bibfield{author}{\bibinfo{person}{Ting-Yun Hsiao} {et~al\mbox{.}}} \bibinfo{year}{2019}\natexlab{}.
\newblock \showarticletitle{{Filter-based Deep-compression with Global Average Pooling for Convolutional Networks}}.
\newblock \bibinfo{journal}{\emph{Journal of Systems Architecture}}  \bibinfo{volume}{95} (\bibinfo{year}{2019}), \bibinfo{pages}{9--18}.
\newblock


\bibitem[\protect\citeauthoryear{Hunter}{Hunter}{2007}]%
        {matplotlib}
\bibfield{author}{\bibinfo{person}{J.~D. Hunter}.} \bibinfo{year}{2007}\natexlab{}.
\newblock \showarticletitle{Matplotlib: A 2D graphics environment}.
\newblock \bibinfo{journal}{\emph{Computing in Science \& Engineering}} \bibinfo{volume}{9}, \bibinfo{number}{3} (\bibinfo{year}{2007}), \bibinfo{pages}{90--95}.
\newblock
\urldef\tempurl%
\url{https://doi.org/10.1109/MCSE.2007.55}
\showDOI{\tempurl}


\bibitem[\protect\citeauthoryear{Inc.}{Inc.}{2024}]%
        {streamlit}
\bibfield{author}{\bibinfo{person}{Streamlit Inc.}} \bibinfo{year}{2024}\natexlab{}.
\newblock \bibinfo{title}{Streamlit: The fastest way to build and share data apps}.
\newblock
\newblock
\urldef\tempurl%
\url{https://streamlit.io/}
\showURL{%
\tempurl}


\bibitem[\protect\citeauthoryear{Jain, Patel, Nagalapatti, Gupta, Mehta, Guttula, Mujumdar, Afzal, Sharma~Mittal, and Munigala}{Jain et~al\mbox{.}}{2020}]%
        {jain2020overview}
\bibfield{author}{\bibinfo{person}{Abhinav Jain}, \bibinfo{person}{Hima Patel}, \bibinfo{person}{Lokesh Nagalapatti}, \bibinfo{person}{Nitin Gupta}, \bibinfo{person}{Sameep Mehta}, \bibinfo{person}{Shanmukha Guttula}, \bibinfo{person}{Shashank Mujumdar}, \bibinfo{person}{Shazia Afzal}, \bibinfo{person}{Ruhi Sharma~Mittal}, {and} \bibinfo{person}{Vitobha Munigala}.} \bibinfo{year}{2020}\natexlab{}.
\newblock \showarticletitle{Overview and importance of data quality for machine learning tasks}. In \bibinfo{booktitle}{\emph{Proceedings of the 26th ACM SIGKDD international conference on knowledge discovery \& data mining}}. \bibinfo{pages}{3561--3562}.
\newblock


\bibitem[\protect\citeauthoryear{Jakubik, V{\"o}ssing, K{\"u}hl, Walk, and Satzger}{Jakubik et~al\mbox{.}}{2024}]%
        {jakubik2024data}
\bibfield{author}{\bibinfo{person}{Johannes Jakubik}, \bibinfo{person}{Michael V{\"o}ssing}, \bibinfo{person}{Niklas K{\"u}hl}, \bibinfo{person}{Jannis Walk}, {and} \bibinfo{person}{Gerhard Satzger}.} \bibinfo{year}{2024}\natexlab{}.
\newblock \showarticletitle{Data-centric artificial intelligence}.
\newblock \bibinfo{journal}{\emph{Business \& Information Systems Engineering}} \bibinfo{volume}{66}, \bibinfo{number}{4} (\bibinfo{year}{2024}), \bibinfo{pages}{507--515}.
\newblock


\bibitem[\protect\citeauthoryear{Keogh, Xi, Wei, and Ratanamahatana}{Keogh et~al\mbox{.}}{2006}]%
        {ucr}
\bibfield{author}{\bibinfo{person}{Eamonn Keogh}, \bibinfo{person}{Xiaopeng Xi}, \bibinfo{person}{Li Wei}, {and} \bibinfo{person}{Chotirat~Ann Ratanamahatana}.} \bibinfo{year}{2006}\natexlab{}.
\newblock \showarticletitle{The UCR time series classification/clustering homepage}.
\newblock \bibinfo{journal}{\emph{URL= http://www. cs. ucr. edu/\~{} eamonn/time\_series\_data}} (\bibinfo{year}{2006}).
\newblock


\bibitem[\protect\citeauthoryear{Khan, ten Thij, and Wilbik}{Khan et~al\mbox{.}}{2025}]%
        {khan2025vertical}
\bibfield{author}{\bibinfo{person}{Afsana Khan}, \bibinfo{person}{Marijn ten Thij}, {and} \bibinfo{person}{Anna Wilbik}.} \bibinfo{year}{2025}\natexlab{}.
\newblock \showarticletitle{Vertical federated learning: A structured literature review}.
\newblock \bibinfo{journal}{\emph{Knowledge and Information Systems}} (\bibinfo{year}{2025}), \bibinfo{pages}{1--39}.
\newblock


\bibitem[\protect\citeauthoryear{Knight}{Knight}{2020}]%
        {knight2020ai}
\bibfield{author}{\bibinfo{person}{Will Knight}.} \bibinfo{year}{2020}\natexlab{}.
\newblock \showarticletitle{{AI can do great things - if it doesn’t burn the planet}}.
\newblock \bibinfo{journal}{\emph{Wired Magazine}} (\bibinfo{year}{2020}).
\newblock


\bibitem[\protect\citeauthoryear{Konstantinou and Paton}{Konstantinou and Paton}{2020}]%
        {konstantinou2020feedback}
\bibfield{author}{\bibinfo{person}{Nikolaos Konstantinou} {and} \bibinfo{person}{Norman~W Paton}.} \bibinfo{year}{2020}\natexlab{}.
\newblock \showarticletitle{{Feedback Driven Improvement of Data Preparation Pipelines}}.
\newblock \bibinfo{journal}{\emph{Information Systems}}  \bibinfo{volume}{92} (\bibinfo{year}{2020}), \bibinfo{pages}{101480}.
\newblock


\bibitem[\protect\citeauthoryear{Li, Diao, Chen, and He}{Li et~al\mbox{.}}{2022}]%
        {9835537}
\bibfield{author}{\bibinfo{person}{Qinbin Li}, \bibinfo{person}{Yiqun Diao}, \bibinfo{person}{Quan Chen}, {and} \bibinfo{person}{Bingsheng He}.} \bibinfo{year}{2022}\natexlab{}.
\newblock \showarticletitle{Federated Learning on Non-IID Data Silos: An Experimental Study}. In \bibinfo{booktitle}{\emph{2022 IEEE 38th International Conference on Data Engineering (ICDE)}}. \bibinfo{pages}{965--978}.
\newblock
\urldef\tempurl%
\url{https://doi.org/10.1109/ICDE53745.2022.00077}
\showDOI{\tempurl}


\bibitem[\protect\citeauthoryear{Li, Wen, Wu, Hu, Wang, Li, Liu, and He}{Li et~al\mbox{.}}{2023}]%
        {9599369}
\bibfield{author}{\bibinfo{person}{Qinbin Li}, \bibinfo{person}{Zeyi Wen}, \bibinfo{person}{Zhaomin Wu}, \bibinfo{person}{Sixu Hu}, \bibinfo{person}{Naibo Wang}, \bibinfo{person}{Yuan Li}, \bibinfo{person}{Xu Liu}, {and} \bibinfo{person}{Bingsheng He}.} \bibinfo{year}{2023}\natexlab{}.
\newblock \showarticletitle{A Survey on Federated Learning Systems: Vision, Hype and Reality for Data Privacy and Protection}.
\newblock \bibinfo{journal}{\emph{IEEE Transactions on Knowledge and Data Engineering}} \bibinfo{volume}{35}, \bibinfo{number}{4} (\bibinfo{year}{2023}), \bibinfo{pages}{3347--3366}.
\newblock
\urldef\tempurl%
\url{https://doi.org/10.1109/TKDE.2021.3124599}
\showDOI{\tempurl}


\bibitem[\protect\citeauthoryear{Liu, Kang, Zou, Pu, He, Ye, Ouyang, Zhang, and Yang}{Liu et~al\mbox{.}}{2024}]%
        {10415268}
\bibfield{author}{\bibinfo{person}{Yang Liu}, \bibinfo{person}{Yan Kang}, \bibinfo{person}{Tianyuan Zou}, \bibinfo{person}{Yanhong Pu}, \bibinfo{person}{Yuanqin He}, \bibinfo{person}{Xiaozhou Ye}, \bibinfo{person}{Ye Ouyang}, \bibinfo{person}{Ya-Qin Zhang}, {and} \bibinfo{person}{Qiang Yang}.} \bibinfo{year}{2024}\natexlab{}.
\newblock \showarticletitle{Vertical Federated Learning: Concepts, Advances, and Challenges}.
\newblock \bibinfo{journal}{\emph{IEEE Transactions on Knowledge and Data Engineering}} \bibinfo{volume}{36}, \bibinfo{number}{7} (\bibinfo{year}{2024}), \bibinfo{pages}{3615--3634}.
\newblock
\urldef\tempurl%
\url{https://doi.org/10.1109/TKDE.2024.3352628}
\showDOI{\tempurl}


\bibitem[\protect\citeauthoryear{Lucivero}{Lucivero}{2020}]%
        {lucivero2020big}
\bibfield{author}{\bibinfo{person}{Federica Lucivero}.} \bibinfo{year}{2020}\natexlab{}.
\newblock \showarticletitle{{Big data, big waste? A reflection on the environmental sustainability of big data initiatives}}.
\newblock \bibinfo{journal}{\emph{Science and Engineering Ethics}} \bibinfo{volume}{26}, \bibinfo{number}{2} (\bibinfo{year}{2020}), \bibinfo{pages}{1009--1030}.
\newblock


\bibitem[\protect\citeauthoryear{Maccioni and Torlone}{Maccioni and Torlone}{2018}]%
        {maccioni2018kayak}
\bibfield{author}{\bibinfo{person}{Antonio Maccioni} {and} \bibinfo{person}{Riccardo Torlone}.} \bibinfo{year}{2018}\natexlab{}.
\newblock \showarticletitle{{KAYAK: a Framework for just-in-time Data Preparation in a Data Lake}}. In \bibinfo{booktitle}{\emph{Advanced Information Systems Engineering: 30th International Conference, CAiSE 2018, Tallinn, Estonia, June 11-15, 2018, Proceedings 30}}. Springer, \bibinfo{pages}{474--489}.
\newblock


\bibitem[\protect\citeauthoryear{Mao, Yu, Huang, Zhang, and Zhang}{Mao et~al\mbox{.}}{2024}]%
        {mao2024green}
\bibfield{author}{\bibinfo{person}{Yuyi Mao}, \bibinfo{person}{Xianghao Yu}, \bibinfo{person}{Kaibin Huang}, \bibinfo{person}{Ying-Jun~Angela Zhang}, {and} \bibinfo{person}{Jun Zhang}.} \bibinfo{year}{2024}\natexlab{}.
\newblock \showarticletitle{Green edge AI: A contemporary survey}.
\newblock \bibinfo{journal}{\emph{Proc. IEEE}} (\bibinfo{year}{2024}).
\newblock


\bibitem[\protect\citeauthoryear{Miao et~al\mbox{.}}{Miao et~al\mbox{.}}{2023}]%
        {miao2023data}
\bibfield{author}{\bibinfo{person}{Zhuqi Miao} {et~al\mbox{.}}} \bibinfo{year}{2023}\natexlab{}.
\newblock \showarticletitle{{A Data Preparation Framework for Cleaning Electronic Health Records and Assessing Cleaning Outcomes for Secondary Analysis}}.
\newblock \bibinfo{journal}{\emph{Information Systems}}  \bibinfo{volume}{111} (\bibinfo{year}{2023}), \bibinfo{pages}{102130}.
\newblock


\bibitem[\protect\citeauthoryear{pandas~development team}{pandas~development team}{2020}]%
        {pandas}
\bibfield{author}{\bibinfo{person}{The pandas~development team}.} \bibinfo{year}{2020}\natexlab{}.
\newblock \bibinfo{booktitle}{\emph{pandas-dev/pandas: Pandas}}.
\newblock
\urldef\tempurl%
\url{https://doi.org/10.5281/zenodo.3509134}
\showDOI{\tempurl}


\bibitem[\protect\citeauthoryear{Pedregosa, Varoquaux, Gramfort, Michel, Thirion, Grisel, Blondel, Prettenhofer, Weiss, Dubourg, et~al\mbox{.}}{Pedregosa et~al\mbox{.}}{2011}]%
        {scikit}
\bibfield{author}{\bibinfo{person}{Fabian Pedregosa}, \bibinfo{person}{Ga{\"e}l Varoquaux}, \bibinfo{person}{Alexandre Gramfort}, \bibinfo{person}{Vincent Michel}, \bibinfo{person}{Bertrand Thirion}, \bibinfo{person}{Olivier Grisel}, \bibinfo{person}{Mathieu Blondel}, \bibinfo{person}{Peter Prettenhofer}, \bibinfo{person}{Ron Weiss}, \bibinfo{person}{Vincent Dubourg}, {et~al\mbox{.}}} \bibinfo{year}{2011}\natexlab{}.
\newblock \showarticletitle{Scikit-learn: Machine learning in Python}.
\newblock \bibinfo{journal}{\emph{Journal of machine learning research}} \bibinfo{volume}{12}, \bibinfo{number}{Oct} (\bibinfo{year}{2011}), \bibinfo{pages}{2825--2830}.
\newblock


\bibitem[\protect\citeauthoryear{Pilgrim and Willison}{Pilgrim and Willison}{2009}]%
        {python}
\bibfield{author}{\bibinfo{person}{Mark Pilgrim} {and} \bibinfo{person}{Simon Willison}.} \bibinfo{year}{2009}\natexlab{}.
\newblock \bibinfo{booktitle}{\emph{Dive Into Python 3}}. Vol.~\bibinfo{volume}{2}.
\newblock \bibinfo{publisher}{Springer}.
\newblock


\bibitem[\protect\citeauthoryear{Polyzotis, Roy, Whang, and Zinkevich}{Polyzotis et~al\mbox{.}}{2018}]%
        {polyzotis2018data}
\bibfield{author}{\bibinfo{person}{Neoklis Polyzotis}, \bibinfo{person}{Sudip Roy}, \bibinfo{person}{Steven~Euijong Whang}, {and} \bibinfo{person}{Martin Zinkevich}.} \bibinfo{year}{2018}\natexlab{}.
\newblock \showarticletitle{Data lifecycle challenges in production machine learning: a survey}.
\newblock \bibinfo{journal}{\emph{ACM Sigmod Record}} \bibinfo{volume}{47}, \bibinfo{number}{2} (\bibinfo{year}{2018}), \bibinfo{pages}{17--28}.
\newblock


\bibitem[\protect\citeauthoryear{Qiu, Parcollet, Fernandez-Marques, Gusmao, Gao, Beutel, Topal, Mathur, and Lane}{Qiu et~al\mbox{.}}{2023}]%
        {qiu2023first}
\bibfield{author}{\bibinfo{person}{Xinchi Qiu}, \bibinfo{person}{Titouan Parcollet}, \bibinfo{person}{Javier Fernandez-Marques}, \bibinfo{person}{Pedro~PB Gusmao}, \bibinfo{person}{Yan Gao}, \bibinfo{person}{Daniel~J Beutel}, \bibinfo{person}{Taner Topal}, \bibinfo{person}{Akhil Mathur}, {and} \bibinfo{person}{Nicholas~D Lane}.} \bibinfo{year}{2023}\natexlab{}.
\newblock \showarticletitle{A first look into the carbon footprint of federated learning}.
\newblock \bibinfo{journal}{\emph{Journal of Machine Learning Research}} \bibinfo{volume}{24}, \bibinfo{number}{129} (\bibinfo{year}{2023}), \bibinfo{pages}{1--23}.
\newblock


\bibitem[\protect\citeauthoryear{Rolnick et~al\mbox{.}}{Rolnick et~al\mbox{.}}{2022}]%
        {rolnick2022tackling}
\bibfield{author}{\bibinfo{person}{David Rolnick} {et~al\mbox{.}}} \bibinfo{year}{2022}\natexlab{}.
\newblock \showarticletitle{{Tackling Climate Change with Machine Learning}}.
\newblock \bibinfo{journal}{\emph{ACM Computing Surveys (CSUR)}} \bibinfo{volume}{55}, \bibinfo{number}{2} (\bibinfo{year}{2022}), \bibinfo{pages}{1--96}.
\newblock


\bibitem[\protect\citeauthoryear{Savazzi, Rampa, Kianoush, and Bennis}{Savazzi et~al\mbox{.}}{2022}]%
        {savazzi2022energy}
\bibfield{author}{\bibinfo{person}{Stefano Savazzi}, \bibinfo{person}{Vittorio Rampa}, \bibinfo{person}{Sanaz Kianoush}, {and} \bibinfo{person}{Mehdi Bennis}.} \bibinfo{year}{2022}\natexlab{}.
\newblock \showarticletitle{An energy and carbon footprint analysis of distributed and federated learning}.
\newblock \bibinfo{journal}{\emph{IEEE Transactions on Green Communications and Networking}} \bibinfo{volume}{7}, \bibinfo{number}{1} (\bibinfo{year}{2022}), \bibinfo{pages}{248--264}.
\newblock


\bibitem[\protect\citeauthoryear{Schwartz et~al\mbox{.}}{Schwartz et~al\mbox{.}}{2020}]%
        {Schwartz2020}
\bibfield{author}{\bibinfo{person}{Roy Schwartz} {et~al\mbox{.}}} \bibinfo{year}{2020}\natexlab{}.
\newblock \showarticletitle{{Green AI}}.
\newblock \bibinfo{journal}{\emph{Commun. ACM}} \bibinfo{volume}{63}, \bibinfo{number}{12} (\bibinfo{year}{2020}), \bibinfo{pages}{54--63}.
\newblock


\bibitem[\protect\citeauthoryear{Shin et~al\mbox{.}}{Shin et~al\mbox{.}}{2020}]%
        {shin2020practical}
\bibfield{author}{\bibinfo{person}{Yunjung Shin} {et~al\mbox{.}}} \bibinfo{year}{2020}\natexlab{}.
\newblock \showarticletitle{{Practical Methods of Image Data Preprocessing for Enhancing the Performance of Deep Learning Based Road Crack Detection}}.
\newblock \bibinfo{journal}{\emph{ICIC Express Letters, Part B: Applications}} \bibinfo{volume}{11}, \bibinfo{number}{4} (\bibinfo{year}{2020}), \bibinfo{pages}{373--379}.
\newblock


\bibitem[\protect\citeauthoryear{Strubell, Ganesh, and McCallum}{Strubell et~al\mbox{.}}{2020}]%
        {Strubell2019}
\bibfield{author}{\bibinfo{person}{Emma Strubell}, \bibinfo{person}{Ananya Ganesh}, {and} \bibinfo{person}{Andrew McCallum}.} \bibinfo{year}{2020}\natexlab{}.
\newblock \showarticletitle{Energy and policy considerations for modern deep learning research}. In \bibinfo{booktitle}{\emph{Proceedings of the AAAI conference on artificial intelligence}}, Vol.~\bibinfo{volume}{34}. \bibinfo{pages}{13693--13696}.
\newblock


\bibitem[\protect\citeauthoryear{Sun et~al\mbox{.}}{Sun et~al\mbox{.}}{2017}]%
        {unreffectiveness}
\bibfield{author}{\bibinfo{person}{Chen Sun} {et~al\mbox{.}}} \bibinfo{year}{2017}\natexlab{}.
\newblock \showarticletitle{{Revisiting Unreasonable Effectiveness of Data in Deep Learning Era}}. In \bibinfo{booktitle}{\emph{Proceedings of the IEEE International Conference on Computer Vision}}. \bibinfo{pages}{843--852}.
\newblock


\bibitem[\protect\citeauthoryear{Thakur, Guzzo, Fortino, and Piccialli}{Thakur et~al\mbox{.}}{2025}]%
        {thakur2025green}
\bibfield{author}{\bibinfo{person}{Dipanwita Thakur}, \bibinfo{person}{Antonella Guzzo}, \bibinfo{person}{Giancarlo Fortino}, {and} \bibinfo{person}{Francesco Piccialli}.} \bibinfo{year}{2025}\natexlab{}.
\newblock \showarticletitle{Green Federated Learning: A new era of Green Aware AI}.
\newblock \bibinfo{journal}{\emph{Comput. Surveys}} (\bibinfo{year}{2025}).
\newblock


\bibitem[\protect\citeauthoryear{Verdecchia, Cruz, Sallou, Lin, Wickenden, and Hotellier}{Verdecchia et~al\mbox{.}}{2022}]%
        {Verdecchia_2022}
\bibfield{author}{\bibinfo{person}{Roberto Verdecchia}, \bibinfo{person}{Luis Cruz}, \bibinfo{person}{June Sallou}, \bibinfo{person}{Michelle Lin}, \bibinfo{person}{James Wickenden}, {and} \bibinfo{person}{Estelle Hotellier}.} \bibinfo{year}{2022}\natexlab{}.
\newblock \showarticletitle{Data-Centric Green AI An Exploratory Empirical Study}. In \bibinfo{booktitle}{\emph{2022 International Conference on ICT for Sustainability (ICT4S)}}. \bibinfo{publisher}{IEEE}, \bibinfo{pages}{35–45}.
\newblock
\urldef\tempurl%
\url{https://doi.org/10.1109/ict4s55073.2022.00015}
\showDOI{\tempurl}


\bibitem[\protect\citeauthoryear{Wang, Tong, Shi, and Xu}{Wang et~al\mbox{.}}{2021}]%
        {9458704}
\bibfield{author}{\bibinfo{person}{Yansheng Wang}, \bibinfo{person}{Yongxin Tong}, \bibinfo{person}{Dingyuan Shi}, {and} \bibinfo{person}{Ke Xu}.} \bibinfo{year}{2021}\natexlab{}.
\newblock \showarticletitle{An Efficient Approach for Cross-Silo Federated Learning to Rank}. In \bibinfo{booktitle}{\emph{2021 IEEE 37th International Conference on Data Engineering (ICDE)}}. \bibinfo{pages}{1128--1139}.
\newblock
\urldef\tempurl%
\url{https://doi.org/10.1109/ICDE51399.2021.00102}
\showDOI{\tempurl}


\bibitem[\protect\citeauthoryear{Werner~de Vargas et~al\mbox{.}}{Werner~de Vargas et~al\mbox{.}}{2023}]%
        {werner2023imbalanced}
\bibfield{author}{\bibinfo{person}{Vitor Werner~de Vargas} {et~al\mbox{.}}} \bibinfo{year}{2023}\natexlab{}.
\newblock \showarticletitle{{Imbalanced Data Preprocessing Techniques for Machine Learning: a Systematic Mapping Study}}.
\newblock \bibinfo{journal}{\emph{Knowledge and Information Systems}} \bibinfo{volume}{65}, \bibinfo{number}{1} (\bibinfo{year}{2023}), \bibinfo{pages}{31--57}.
\newblock


\bibitem[\protect\citeauthoryear{Whang, Roh, Song, and Lee}{Whang et~al\mbox{.}}{2023}]%
        {whang2023data}
\bibfield{author}{\bibinfo{person}{Steven~Euijong Whang}, \bibinfo{person}{Yuji Roh}, \bibinfo{person}{Hwanjun Song}, {and} \bibinfo{person}{Jae-Gil Lee}.} \bibinfo{year}{2023}\natexlab{}.
\newblock \showarticletitle{Data collection and quality challenges in deep learning: A data-centric ai perspective}.
\newblock \bibinfo{journal}{\emph{The VLDB Journal}} \bibinfo{volume}{32}, \bibinfo{number}{4} (\bibinfo{year}{2023}), \bibinfo{pages}{791--813}.
\newblock


\bibitem[\protect\citeauthoryear{Wiesner, Khalili, Grinwald, Agrawal, Thamsen, and Kao}{Wiesner et~al\mbox{.}}{2024}]%
        {wiesner2024fedzero}
\bibfield{author}{\bibinfo{person}{Philipp Wiesner}, \bibinfo{person}{Ramin Khalili}, \bibinfo{person}{Dennis Grinwald}, \bibinfo{person}{Pratik Agrawal}, \bibinfo{person}{Lauritz Thamsen}, {and} \bibinfo{person}{Odej Kao}.} \bibinfo{year}{2024}\natexlab{}.
\newblock \showarticletitle{Fedzero: Leveraging renewable excess energy in federated learning}. In \bibinfo{booktitle}{\emph{Proceedings of the 15th ACM International Conference on Future and Sustainable Energy Systems}}. \bibinfo{pages}{373--385}.
\newblock


\bibitem[\protect\citeauthoryear{Wu, Xing, Chen, Dinh, Luo, Ooi, Xiao, and Zhang}{Wu et~al\mbox{.}}{2023}]%
        {wu2023falcon}
\bibfield{author}{\bibinfo{person}{Yuncheng Wu}, \bibinfo{person}{Naili Xing}, \bibinfo{person}{Gang Chen}, \bibinfo{person}{Tien Tuan~Anh Dinh}, \bibinfo{person}{Zhaojing Luo}, \bibinfo{person}{Beng~Chin Ooi}, \bibinfo{person}{Xiaokui Xiao}, {and} \bibinfo{person}{Meihui Zhang}.} \bibinfo{year}{2023}\natexlab{}.
\newblock \showarticletitle{Falcon: A privacy-preserving and interpretable vertical federated learning system}.
\newblock \bibinfo{journal}{\emph{Proceedings of the VLDB Endowment}} \bibinfo{volume}{16}, \bibinfo{number}{10} (\bibinfo{year}{2023}), \bibinfo{pages}{2471--2484}.
\newblock


\bibitem[\protect\citeauthoryear{Zha, Bhat, Lai, Yang, Jiang, Zhong, and Hu}{Zha et~al\mbox{.}}{2025}]%
        {zha2025data}
\bibfield{author}{\bibinfo{person}{Daochen Zha}, \bibinfo{person}{Zaid~Pervaiz Bhat}, \bibinfo{person}{Kwei-Herng Lai}, \bibinfo{person}{Fan Yang}, \bibinfo{person}{Zhimeng Jiang}, \bibinfo{person}{Shaochen Zhong}, {and} \bibinfo{person}{Xia Hu}.} \bibinfo{year}{2025}\natexlab{}.
\newblock \showarticletitle{Data-centric artificial intelligence: A survey}.
\newblock \bibinfo{journal}{\emph{Comput. Surveys}} \bibinfo{volume}{57}, \bibinfo{number}{5} (\bibinfo{year}{2025}), \bibinfo{pages}{1--42}.
\newblock


\end{thebibliography}

\end{document}